%% file: paper.tex
\pdfoutput=1

\documentclass[11pt]{article}

\input{header}
\usepackage[T1]{fontenc}

\usepackage[utf8]{inputenc}

\usepackage{microtype}

\usepackage{inconsolata}

\usepackage[capitalize, nameinlink]{cleveref}
\crefformat{section}{\S#2#1#3} %
\crefformat{subsection}{\S#2#1#3}
\crefformat{subsubsection}{\S#2#1#3}
\crefformat{appendix}{App. #2#1#3}

\newenvironment{itemize*}%
  {\begin{itemize}%
    \setlength{\itemsep}{.2pt}%
    \setlength{\parskip}{.2pt}%
    \setlength{\topsep}{.5pt}}%
  {\end{itemize}}

\newcommand*{\figuretitle}[1]{%
    {\centering%
    \textbf{#1}%
    \par\medskip}%
}

\newcommand*{\affaddr}[1]{#1} %
\newcommand*{\affmark}[1][*]{\textsuperscript{#1}}
\newcommand*{\email}[1]{\texttt{#1}}

\newcommand{\out}[1]{${O}{#1}$}

\title{Outlier Dimensions that Disrupt Transformers are Driven by Frequency}

\author{Giovanni Puccetti\affmark[1, 2, 4], Anna Rogers\affmark[3,4], Aleksandr Drozd\affmark[4], Felice Dell'Orletta\affmark[2] \\
\affaddr{\affmark[1] \normalsize Scuola Normale Superiore, Pisa, Italy} \\
\affaddr{\affmark[2] \normalsize Istituto di Linguistica Computazionale ``Antonio Zampolli'', Pisa, ItaliaNLPLab - \emph{www.italianlp.it}} \\
\affaddr{\affmark[3] \normalsize Center for Social Data Science, University of Copenhagen, Denmark} \\
\affaddr{\affmark[4] \normalsize RIKEN Center for Computational Science, Japan} \\
{\normalsize \email{giovanni.puccetti@sns.it}}, \hspace{0.cm}
{\normalsize \email{arogers@sodas.ku.dk}}, \hspace{0.cm}\\
{\normalsize \email{alex@blackbird.pw}}, \hspace{0.cm}
{\normalsize \email{felice.dellorletta@ilc.cnr.it}}, \hspace{0.cm}
}

\begin{document}
\maketitle
\begin{abstract}
While Transformer-based language models are generally very robust to pruning, there is the recently discovered outlier phenomenon: disabling only 48 out of 110M parameters in BERT-base drops its performance by nearly 30\% on MNLI. We replicate the original evidence for the outlier phenomenon and we link it to the geometry of the embedding space. We find that in both BERT and RoBERTa the magnitude of hidden state coefficients corresponding to outlier dimensions correlates with the frequency of encoded tokens in pre-training data, and it also contributes to the ``vertical'' self-attention pattern enabling the model to focus on the special tokens. This explains the drop in performance from disabling the outliers, and it suggests that to decrease anisotropicity in future models we need pre-training schemas that would better take into account the skewed token distributions.

\end{abstract}

\input{chapters/1_introduction}
\input{chapters/2_methodology}
\input{chapters/3_other_transformers}
\input{chapters/4_outlier_analysis}
\input{chapters/5_discussion}

\input{chapters/6_conclusion}

\input{chapters/7_limitations}

\input{chapters/8_broader_impact}

\section*{Acknowledgements}
We would like to thank Olga Kovaleva, Anna Rumshisky, and the anonymous reviewers for their insightful comments. 
This work is partially supported by JST KAKENHI grant JP22H03600 and JST CREST grant JPMJCR19F5. This work used computational resources of the supercomputer Fugaku provided by RIKEN through the HPCI Fugaku General Access (Small-Scale) Project (Project ID: hp210265).

\bibliography{anthology,custom}
\bibliographystyle{acl_natbib}

\input{chapters/appendix}

\end{document}

%% file: header.tex
\usepackage{EMNLP2022}

\usepackage{times}
\usepackage{latexsym}
\usepackage{booktabs}
\usepackage{xcolor}
\usepackage{adjustbox}
\usepackage{amssymb}
\usepackage{todonotes}
\usepackage{caption}
\usepackage{subcaption}
\usepackage{tabularx}

\newcommand{\split}{\emph{SENTENCE}} %
\newcommand{\onesep}{\emph{CHUNK}} %
\newcommand{\randomize}{\emph{SENTENCE\_FREQ}} %

%% file: chapters/1_introduction.tex
\section{Introduction}
\label{sec:introduction}

The current Transformer-based language models are heavily overparametrized, which explains why it is possible to prune these models by up to 30-40\% \cite[][inter alia]{gordon-etal-2020-compressing, movement-pruning,PrasannaRogersEtAl_2020_When_BERT_Plays_Lottery_All_Tickets_Are_Winning,ChenFrankleEtAl_2020_Lottery_Ticket_Hypothesis_for_Pre-trained_BERT_Networks} without a significant drop in performance. However, it has recently been shown that multiple Transformer-based language models (LMs) are highly sensitive to removal of \emph{outlier dimensions} \cite{KovalevaKulshreshthaEtAl_2021_BERT_Busters_Outlier_Dimensions_that_Disrupt_Transformers}: the parameters (weights and biases) in the output element of a Transformer layer, the magnitude of which is unusually large within the layer (consistently in the same dimension across the model layers). For BERT model family the output element is the LayerNorm, as shown in \autoref{fig:att_is_all}.

Although these parameters constitute less than 0.0001\% of the full BERT \cite{devlin2019bert} model, removing them significantly degrades BERT's performance. %
\citet{PuccettiMiaschiEtAl_2021_How_Do_BERT_Embeddings_Organize_Linguistic_Knowledgea} find that the same parameters are particularly relevant in several linguistic probing tasks. These dimensions affect the vector representation of different tokens in the same way, making the embedding space less isotropic and thus reducing its representational power \cite{LiangCaoEtAl_2021_Learning_to_Remove_Towards_Isotropic_Pre-trained_BERT_Embedding}. Outlier dimensions have also been found to make model quantization challenging \cite{bondarenko-etal-2021-understanding,llmint8} as they need to be treated separately from others when defining quantization schemes. Thus there are both conceptual and practical reasons supporting a deeper study of this phenomenon.
\input{figures/att_is_all_you_need_fig}

\begin{table*}[ht]
    \small
    \centering
    \begin{adjustbox}{width=\linewidth}
    \input{tables/bert_base_uncased_full_std}
    \end{adjustbox}
    \caption{Average BERT scores over 5 runs on GLUE benchmarks with the effect of outlier removal. The rows \emph{1 random removed} and \emph{2 random removed} show the average over 5 removals of random non outliers (1 or 2 at a time respectively) for 5 different fine-tuned models}
    \label{tab:bert-outliers}
\end{table*}

What is not clear at this point is the mechanism behind the emergence of outliers. We replicate the original findings in BERT and RoBERTa, and we contribute new evidence \textbf{directly linking the outlier phenomenon with the frequency of encoded tokens in the pre-training data, as well as the self-attention pattern focusing on special tokens}. We also present evidence for two kinds of outliers: some of them affect the Masked Language Model (MLM) performance the most in the middle layers (where the correlation with token frequency is at its peak), and for others the impact grows towards the final layers (although correlation with token frequency decreases). This work contributes to mechanistic understanding of Transfromer-based LMs, and it might be useful for future research on decreasing anisotropy in pre-trained LMs.

%% file: figures/att_is_all_you_need_fig.tex
\begin{figure}[t]
    \centering
    \includegraphics[scale=0.5]{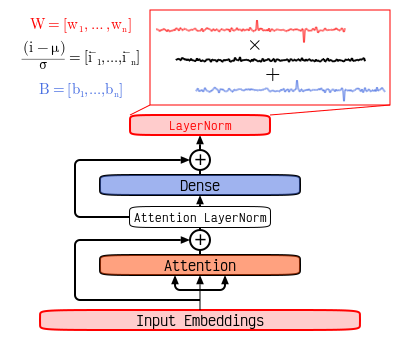}
    \caption{The Transformer Layer architecture diagram with outliers at the normalization layer (LayerNorm).}
    \label{fig:att_is_all}
\end{figure}

%% file: tables/bert_base_uncased_full_std.tex
\begin{tabular}{ccccccccccc}
\toprule
\emph{bert-base-uncased} &          cola &          mnli &       mnli-mm &           mrpc &          qnli &           qqp &           rte &          sst2 &           stsb \\
\midrule
baseline        &  56.9 +/- 1.5 &  84.5 +/- 0.2 &  84.8 +/- 0.4 &   84.3 +/- 1.1 &  91.4 +/- 0.1 &  91.1 +/- 0.1 &  66.3 +/- 1.6 &  92.8 +/- 0.5 &   89.0 +/- 0.3 \\
\midrule
1 random removed &  56.5 +/- 1.5 &  84.5 +/- 0.2 &  84.8 +/- 0.4 &  84.5 +/- 0.9 &  91.3 +/- 0.1 &  91.1 +/- 0.1 &  66.6 +/- 1.7 &  92.8 +/- 0.4 &  89.0 +/- 0.3 \\
w/o 308     &  47.3 +/- 1.2 &  81.4 +/- 1.1 &  82.2 +/- 1.1 &  54.0 +/- 12.3 &  88.9 +/- 0.8 &  89.1 +/- 1.6 &  62.1 +/- 3.2 &  90.8 +/- 1.1 &  56.9 +/- 17.4 \\
w/o 381     &  33.8 +/- 9.4 &  73.2 +/- 2.1 &  73.8 +/- 2.0 &  64.6 +/- 15.2 &  80.3 +/- 1.5 &  79.8 +/- 3.2 &  55.8 +/- 1.6 &  87.9 +/- 1.0 &   78.1 +/- 4.8 \\
\midrule
2 random removed &  56.4 +/- 1.5 &  84.5 +/- 0.2 &  84.8 +/- 0.4 &  84.3 +/- 0.9 &  91.3 +/- 0.1 &  91.1 +/- 0.1 &  66.6 +/- 1.7 &  92.8 +/- 0.5 &  88.9 +/- 0.3 \\
w/o 308 \& 381 &  15.9 +/- 4.2 &  58.4 +/- 3.3 &  59.0 +/- 3.5 &  55.1 +/- 16.3 &  74.5 +/- 1.3 &  74.6 +/- 4.6 &  55.3 +/- 4.5 &  76.0 +/- 2.4 &  35.7 +/- 15.3 \\
\bottomrule
\end{tabular}

%% file: chapters/2_methodology.tex
\section{Methodology}
\label{sec:outlier-def}

According to \citeauthor{KovalevaKulshreshthaEtAl_2021_BERT_Busters_Outlier_Dimensions_that_Disrupt_Transformers}, \textit{outliers} are parameters (both weights and biases) in the final element of a Transformer layer (LayerNorm for BERT family, final MLP for GPT-2), which have unusually high magnitude\footnote{The original definition of outliers is not entirely formal, and needs to be further specified for particular models: the magnitude of the outliers was within 2 standard deviations from the mean for RoBERTa, and within 3 for BERT.} within the layer. \textit{Outlier dimensions} are those dimensions at which outlier parameters are found consistently across the model layers. 

The reason \citeauthor{KovalevaKulshreshthaEtAl_2021_BERT_Busters_Outlier_Dimensions_that_Disrupt_Transformers} study these parameters is that when they are disabled, the model performance on downstream tasks is greatly reduced. Since not all parameters that can be identified by magnitude and position criteria have that effect, we add this property to the definition. In this work the term \textit{outlier dimension} refers to the dimensions with parameters meeting the magnitude criteria across layers and having at least 5x more damaging effect on accuracy on a representative downstream task, for which we choose MNLI (see \cref{sec:repl-prior}). %

To disable the outliers, unless stated otherwise, we set to zero both the LayerNorm \emph{weight} and \emph{bias} parameters for all layers (24 parameters in total for one outlier dimension in BERT and RoBERTa-base)\footnote{Note that this is equivalent to zeroing out the outlier of the hidden state generated by that layer.}. See \cref{app:outlier-list} for the full list of outliers identified for all models in this study.

We use the notation $O$ to refer to specific LayerNorm outlier parameters (in BERT model family): e.g. \out{381} to indicate ``an outlier with index 381''. Since outlier indices are a constant for a given model, in this study we will also discuss \emph{hidden state outlier dimensions}: the coefficient of the hidden state with the same index as the outlier.

We experiment with BERT-base \cite{devlin2019bert} (\emph{"bert-base-uncased"}), RoBERTa-base \cite{liu2019roberta} (\emph{"roberta-base"}) and Vision Transformer \cite{vit} (\emph{"google/vit-base-patch16-224-in21k"}) from the \emph{transformers} library\footnote{\label{footnote:transformers_repo}\url{https://github.com/huggingface/transformers}}. For the experiments on pre-training dynamics we rely on the checkpoints with seed 1 provided by \citet{sellam2022the-multiberts}\footnote{\label{footnote:multiberts_repo}\url{https://github.com/google-research/language/tree/master/language/multiberts}}.

Hardware, implementation and energy expenditure details are outlined in \cref{app:reproducibility}. We release the code to replicate our experiments\footnote{\label{github_link}\url{https://github.com/gpucce/outliersvsfreq/tree/main}}.

%% file: chapters/3_other_transformers.tex
\section{Outliers Phenomenon in Transformers}
\label{sec:outlier-phenomenon}

\subsection{Replicating Prior Evidence}
\label{sec:repl-prior}

We start by replicating \citeauthor{KovalevaKulshreshthaEtAl_2021_BERT_Busters_Outlier_Dimensions_that_Disrupt_Transformers}'s experiments identifying the outliers for BERT- and RoBERTa-base (\out{308} and \out{381}, \out{77} and \out{588} respectively), and their effect on downstream task performance. 

\cref{tab:bert-outliers} shows the average performance and standard deviation of BERT-base over 5 fine-tuning runs for eight GLUE\footnote{We consider 8 GLUE \cite{wang-etal-2018-glue} tasks: CoLA \cite{cola_ds}, SST \cite{sst_ds}, MRPC \cite{mrpc_ds}, STSB \cite{stsb_ds}, MNLI \cite{mnli_ds}, QNLI \cite{qnli_ds} and RTE \cite{rte_ds}. We exclude WNLI task, which BERT is unable to ``learn'' \cite{PrasannaRogersEtAl_2020_When_BERT_Plays_Lottery_All_Tickets_Are_Winning}.} tasks. Thus we successfully replicate the original experiment on model degradation after\footnote{\citet{KovalevaKulshreshthaEtAl_2021_BERT_Busters_Outlier_Dimensions_that_Disrupt_Transformers} also show that if the outliers are removed \textit{before} fine-tuning, the model is able to recover without any negative effects.} removal of the outliers. Since the effect is consistent across GLUE tasks,
we use MNLI as a representative downstream task in the remaining experiments. We also confirm that RoBERTa-base behaves similarly (see \cref{app:roberta}).

\subsection{Outliers in Other Transformers}
\label{sec:other_transformers}

\citet{KovalevaKulshreshthaEtAl_2021_BERT_Busters_Outlier_Dimensions_that_Disrupt_Transformers} focus exclusively on Transformer-based LMs. To establish whether outliers could be something specific to pre-training on language data, we investigate the presence of outliers in the Vision Transformer (ViT) \cite{vit}. \Cref{tab:vit_outliers} shows ViT accuracy on CIFAR10 \cite{cifar10} and CIFAR100 \cite{Krizhevsky09learningmultiple}: image classification tasks with a choice between 10 and 100 possible labels respectively. Using the magnitude and position criteria we identify candidates \out{759} and \out{187}, and we experiment with disabling one or both of them, as well as randomly selected dimensions as a control. For this model, the accuracy on MNLI can't be used as a measure for outliers, instead we use the accuracy on CIFAR100.

\begin{table}[!t]
    \centering
    \begin{adjustbox}{width=.8\linewidth}
    \input{tables/vit_outliers_cifar}
    \end{adjustbox}
    \caption{Outlier removal effect for Visual Transfromer.}%
    \label{tab:vit_outliers}
\end{table}

We see that, for CIFAR100, with both outliers disabled the model experiences $\approx7\%$ loss in accuracy, but that does not happen for CIFAR10. The reason for that could be that CIFAR10 is a much simpler task, on which the model achieves above 98.5\% accuracy. If the model succeeds in positioning the small number of classes sufficiently far apart in the representation space, then even the loss of outliers might be insufficient to disrupt that. If that is the reason for discrepancy between CIFAR10 and CIFAR100, then perhaps the 100-class classification is still an easier problem than the GLUE tasks, for which BERT degrades in performance significantly more (see \cref{tab:roberta-outliers-full-glue}).

We also explored two other Transformer-based models: ESM trained on protein sequences \cite{protein-transformer} and Wav2Vec trained on audio data \cite{wav2vec}. We found no evidence for outliers there. This could be due to the fact that both of these models have a very small ``vocabulary'' (30-40 ``tokens'' vs tens of thousands for LMs).

\subsection{Emergence of Outliers in Pre-Training}

\citet{KovalevaKulshreshthaEtAl_2021_BERT_Busters_Outlier_Dimensions_that_Disrupt_Transformers} pre-train a BERT-medium model for up to 250,000 steps. They find that outliers emerge relatively early in pre-training (step 50,000), and at about the same time LM perplexity starts to improve. A limitation of this experiment is a relatively small model, and the fact that both observed events coincide with the warm-up ending. 

We examine the full BERT-base checkpoints released by \citet{sellam2022the-multiberts}, who pre-train five models from scratch with different random initializations. For each model they release the checkpoints for every 20,000 steps between 0 and 200,000 steps, and after that -- for every 100,000 steps up to 2,000,000. We use the seed numbered as 1 (zero indexed). Like BERT-base and RoBERTa-base (\cref{sec:outlier-def}), we find that this BERT also has two outliers, \out{218} and \out{674}, the same for all the checkpoints for this seed.

We investigate the main outlier effect: the drop in performance of the model fine-tuned on our chosen representative downstream task, MNLI-matched \cite{mnli}.  \Cref{fig:multiberts_seed_1_mnli_acc_alone} shows the accuracy for all the checkpoints from seed 1, comparing the full model with the model with \out{218}, \out{674}, and both \out{218} and \out{674} removed. The expected effect is clearly observed after step 80000 for \out{218} and \out{218} + \out{674}, but not \out{674} alone. This is consistent with the findings of \citet{KovalevaKulshreshthaEtAl_2021_BERT_Busters_Outlier_Dimensions_that_Disrupt_Transformers} who also report various size of effects for outliers identified purely by magnitude. The results for MNLI-mismatched are similar and available in \cref{app:pre-training}.

\input{figures/multiberts_over_pretraining/multiberts_mnli_matched_over_pretraining}

After step 80,000 the full model steadily increases in accuracy, reaching 83.5\% at step $10^6$. Training for $10^6$ more steps only achieves $\approx1\%$ gain, illustrating the diminishing returns effect with further pre-training. The performance without outliers degrades over time, but at the later stages of pre-training (not observed by \citeauthor{KovalevaKulshreshthaEtAl_2021_BERT_Busters_Outlier_Dimensions_that_Disrupt_Transformers}) that trend is not steady: after $\approx10^6$ steps the model accuracy with either \out{218} or \out{218} + \out{674} removed slowly grows over time, often with high variance between the ``neighboring'' checkpoints.

Another observation from \cref{fig:multiberts_seed_1_mnli_acc_alone} is that after the first $10^6$ steps\footnote{Interestingly, the number of $10^6$ steps is also the number of training steps mentioned in the original BERT paper \cite{devlin2019bert}, and even the models by \citet{sellam2022the-multiberts} (also from Google) do not match the originally reported performance at the original amount of pre-training. \citet{sellam2022the-multiberts} state that they need to train for twice longer to reach comparable performance on all the tasks from GLUE \cite{wang-etal-2018-glue} and SQuAD \cite{rajpurkar-etal-2016-squad}.}
, the difference between the accuracy of the model without the most disrupting outlier \out{218} and \out{674} increases. This suggests that the dynamics for the two outliers are different: while one gains importance from the early stages of pre-training, the other one rises after more optimization steps\footnote{\Cref{fig:multiberts_seed_1_mnli_acc_alone} shows the accuracy with different classification heads initialization. See \cref{app:pre-training} for the similar case with fixed initialization.}. This may be related to the different behavior for the hidden state dimensions corresponding to the two outliers, which we will present in \cref{sec:outliers-effect}.

%% file: tables/vit_outliers_cifar.tex
\begin{tabular}{lcc}
\toprule
\emph{Outliers removed} & CIFAR10 & CIFAR100 \\
\midrule
Full model & 98.6 & 92.5 \\
\midrule
1 random dimension & 98.6 & 92.5 \\
\out{759} & 98.6 & 92.3 \\
\out{187} & 98.6 & 90.5 \\
\midrule
2 random dimensions & 98.6 & 92.4 \\
\out{759} + \out{187} & 98.5 & \textbf{84.9} \\
\bottomrule
\end{tabular}

%% file: figures/multiberts_over_pretraining/multiberts_mnli_matched_over_pretraining.tex
\begin{figure}[tbp]
    \centering
    \includegraphics[width = \linewidth,keepaspectratio,trim=0 0 0 45,clip]{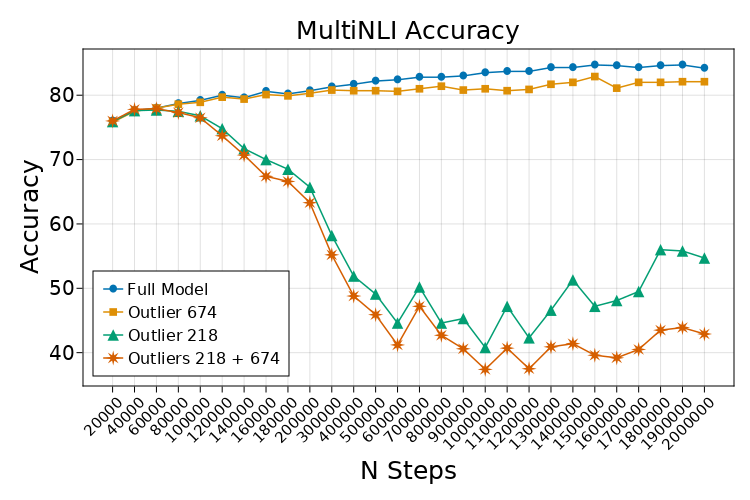}
    \caption{
        The accuracy on MNLI-matched of the checkpoints for BERT-base (seed 1) by \citet{sellam2022the-multiberts} for full model or with each outlier removed. 
    }
    \label{fig:multiberts_seed_1_mnli_acc_alone}
\end{figure}

%% file: chapters/4_outlier_analysis.tex
\section{What Do the Outliers Impact?}
\label{sec:outliers-effect}

\subsection{Effects on Masked Language Modeling}
\label{sec:mlm-effect}

So far we know that disabling the outliers negatively affects BERT downstream task performance (\cref{fig:multiberts_seed_1_mnli_acc_alone}), but it is unclear \textit{why} that happens. Since LMs rely on statistical patterns of token co-occurrence, token frequency in pre-training data\footnote{To estimate the frequency in the pre-training data we use a corpus similar to BERT pre-training data: it contains the Book Corpus \cite{bookcorpus} and Wikipedia dump from November 1st 2021.} could be expected to affect the learned representations. We investigate whether outlier removal affects what kinds of tokens (in terms of their frequency in pre-training data) the MLM predicts.

\Cref{fig:bert_generated_tokens} shows the frequency of tokens predicted by the model over $200,000$ sentences from Wikipedia. We use the standard masking strategy: 15\% tokens masked randomly.
For BERT-base we observe that \textbf{the model with disabled outliers consistently predicts more tokens that were highly frequent in the training data}, and fewer tokens that were rare. RoBERTa shows a similar behaviour (see \cref{app:roberta} for the details).

We also considered if the outliers impact the distribution of POS tags of the predicted tokens. We found that disabling \out{381} is the most disruptive and that, similarly to \out{308}, it pushes the model towards predicting more nouns, punctuation, symbols and adpositions (see \cref{app:outliers-pos-shift} for details).

\subsection{Token Frequency Vs Performance}
\label{sec:influence_of_frequency}

If outlier removal impacts the MLM ability to \textit{predict} tokens it observed less often in pre-training  (\cref{sec:mlm-effect}), could it also impact %
the model \textit{encoding} of tokens more/less frequently seen in pre-training? 

\input{figures/bert-base-uncased/broken_generated_frequencies}

The LayerNorm outliers are an intrinsic property of the model itself. For this experiment we need to consider the interaction between the model and its input data. Hence we consider \textit{the hidden state outlier dimensions}: the hidden state parameters at the dimensions corresponding to the outlier dimensions. They are the most affected by the outlier removal, since zeroing out a LayerNorm parameter removes precisely this component.

In this experiment we encode the validation set of Wikitext-v2 \cite{wikitext} by BERT-base, and we measure the Pearson correlation between pre-training data frequency of encoded tokens, and the magnitude of the hidden state parameters corresponding to the outlier dimensions (\out{308} and \out{381}) in each layer (see \cref{app:reproducibility} for more details). The results are presented in \cref{fig:pre_trained_outlier_corrtofreq_bert-base-uncased}.
We also track across all layers the main outlier effect (performance degradation when the outliers are disabled) in MLM and MNLI tasks, as shown in \cref{fig:bert-base-uncased_nli_compared_to_mlm}. 

We find that for the hidden state parameters corresponding to both \out{308} and \out{381} the correlation between their magnitude and encoded token frequency is much higher than for random dimensions, but they exhibit different layer-wise trends for that correlation vs impact on model performance:

\input{figures/bert-base-uncased/pre_trained_bert-base-uncased_correlation_with_frequency}

\paragraph{Case 1: the magnitude of the hidden state parameters corresponding to the outlier dimensions is directly proportional to both its correlation with the encoded token frequency, and performance drop after removal of LayerNorm outlier parameters.} For the hidden state dimension corresponding to \out{381}, the correlation of the hidden state parameter magnitude with the encoded token frequency is closer to zero at the initial and final layers, and high in the middle layers (this trend continues until layer 9 when special tokens are included). 
\cref{fig:bert-base-uncased_compared_mlm} shows that the removal of \out{381} has largest impact (in both MNLI and MLM) in layers 4-6. 
Coincidentally, \cref{fig:pre_trained_outlier_corrtofreq_bert-base-uncased_wo_special_tok} shows that layers 4-6 are also the layers where the magnitude of hidden state dimension corresponding to \out{381} correlates with token frequency the most.

\paragraph{Case 2: the magnitude of the hidden state parameters corresponding to the outlier dimensions, and their correlation with the encoded token frequency are both inversely proportional to the performance drop after removal of LayerNorm outlier parameters.} For \out{308} the pattern is the opposite: the magnitude of its corresponding hidden state parameter strongly correlates with encoded token frequency at the initial layers, but not in the final ones. However, \cref{fig:bert-base-uncased_compared_mlm} shows that the removal of this LayerNorm outlier has a larger impact on MLM loss on the final layers\footnote{The main discrepancy in this pattern is the frequency correlation of the hidden state dimension corresponding to \out{308}, and its MLM loss at the last layer. However, the lower loss can be a consequence of the parameter not affecting any following Transformer layer.}. As a result, the removal of \out{308} is less harmful for most downstream tasks as shown in \cref{tab:bert-outliers} because fine-tuning mostly affects the layers closer to the output \cite{LiuGardnerEtAl_2019_Linguistic_Knowledge_and_Transferability_of_Contextual_Representations,kovaleva-etal-2019-revealing}, therefore it cancels a part of the effect of disabling this parameter.

To confirm that this is not a pattern specific to BERT we also perform the same experiments for RoBERTa-base, and we find that it also has the two kinds of outliers with the direct and inverse relationship to performance drop (\out{588} and \out{77} respectively). The data for these experiments is available in \cref{app:roberta}. 

\input{figures/bert-base-uncased/bert-base-uncased_nli_compared_to_mlm}

Since BERT encodes sequences always starting with `[CLS]' and ending with `[SEP]', these special tokens could store positional information, and they are also highly frequent. Therefore we repeat the experiment discarding them (\cref{fig:pre_trained_outlier_corrtofreq_bert-base-uncased_wo_special_tok}), but the overall trend is not affected.

\input{figures/bert-base-uncased/bert-base-uncased_attention_sample.tex}

\subsection{What Happens to Attention?}
\label{sec:attention}
In \cref{sec:influence_of_frequency} we showed that there is a correlation between the magnitude of the hidden state parameters corresponding to outlier dimensions, and the token frequency in the pre-training data.
Prior work \cite{clark-etal-2019-bert,kovaleva-etal-2019-revealing} showed that BERT self-attention often ``points'' to highly frequent tokens, including the special tokens and punctuation marks. Given this, our next question is whether the outliers also affect the self-attention patterns. As argued by \citet{pmlr-v139-dong21a}, attention alone would map tokens to very low dimensional spaces, and in that case the outlier phenomenon would be consistent with such a mapping.

We find that there is indeed such an effect. To illustrate it we encode a MNLI sample with BERT-base. \cref{fig:bert_qualitative_att_plot} shows the self-attention maps for the 12 heads of the 10th layer\footnote{We choose the 10th layer because prior work suggests that the layers closer to the output are more affected by fine-tuning \cite{kovaleva-etal-2019-revealing}, and also encode more task-specific information \cite{LiuGardnerEtAl_2019_Linguistic_Knowledge_and_Transferability_of_Contextual_Representations}.}, directly comparing the self-attention in a full model vs a model with the outlier dimensions removed.

The most conspicuous difference is the fact that the vertical bars in the self-attention maps of the full model vanish once the outliers are zeroed out. This ``vertical'' attention pattern has been reported before \cite{kovaleva-etal-2019-revealing}, and in BERT it often corresponds to attention to special tokens and punctuation. It may seem that without the outliers the diagonal patterns become more salient, but in fact they are also present with the intact outliers, and their increased saliency in the plot is simply an effect of softmax normalization.

\Cref{fig:bert_qualitative_att_plot} only shows a single example. To establish whether this effect is stable, we encode 1500 sequences from Wikitext-v2 validation set and measure the Pearson correlation between \emph{average vertical attention value} of each token (the average over attention columns in the encoded sequence), and the magnitude of the hidden state parameters corresponding to the outlier dimensions. In cases of the ``vertical'' self-attention pattern, the average vertical attention value would be relatively high.

\Cref{fig:pre_trained_bert_corr_with_att} shows the results of this experiment, which we repeat with and without BERT special tokens (`[CLS]' and `[SEP]'). As a control,  \cref{fig:pre_trained_bert_corr_with_att_random_with_spec_toks} and \cref{fig:pre_trained_bert_corr_with_att_random_without_spec_toks} show the average correlation over a sample of hidden state parameters at random dimensions. For the randomly picked weights the correlation is $\approx0$, which is expected (since these vectors have length 768, the individual dimensions of randomly sampled vectors should have a negligible contribution).

\input{figures/bert-base-uncased/pre_trained_bert-base-uncased_correlation_with_attention}

Compared to random dimensions, the hidden state parameters at dimensions corresponding to both \out{308} and \out{381} have on average a significantly higher correlation between their magnitude and average self-attention query values. This confirms that the pattern shown in \cref{fig:bert_qualitative_att_plot} is prevalent, and the tokens with high hidden state outlier dimension value tend to also have high average value over attention columns, i.e. \textbf{they are attended to by most other tokens}.

An unexpected pattern is represented by the negative correlations in \cref{fig:pre_trained_bert_corr_with_att_308_with_spec_toks} and \cref{fig:pre_trained_bert_corr_with_att_308_without_spec_toks} at initial and final layers. We argue that at early layers this happens because the vertical patterns are less frequent, while at the final ones because the outliers in those layers are less relevant. The trend is similar to what we observed in \cref{fig:pre_trained_outlier_corrtofreq_bert-base-uncased}.

We also observe several trends that mirror the observations from \cref{sec:influence_of_frequency}:

\begin{itemize*}
    \item The hidden state parameter value corresponding to \out{308} has a higher correlation with average vertical attention value since the initial layers (except the very first) which decreases at the final layers. For parameters corresponding to \out{381}, the correlation grows at layer 4-5 and then vanishes at the final one. Both of these trends are consistent with \cref{fig:pre_trained_outlier_corrtofreq_bert-base-uncased} showing the correlation to frequency.
    \item Special tokens affect these trends: \cref{fig:pre_trained_bert_corr_with_att_308_without_spec_toks} and \cref{fig:pre_trained_bert_corr_with_att_381_without_spec_toks} show that excluding them does not fundamentally change the pattern, but the results become less stable across heads.
    \item Both \cref{fig:pre_trained_outlier_corrtofreq_bert-base-uncased} and \cref{fig:pre_trained_bert_corr_with_att} show large variation as the information flows through the model, which suggests that the effect is not entirely formed at the model input.
\end{itemize*}

Overall the results of this experiment suggest that \textbf{the relationship between the outlier phenomenon and encoded token frequencies in pre-training data also affects the self-attention mechanism of BERT}. In particular, it affects the ``vertical'' attention pattern in which a token is attended to by most other tokens, and which was previously reported for the high-frequency special tokens. We confirmed that RoBERTa self-attention exhibits a similar pattern (see \cref{app:roberta}).

\subsection{What Causes the Outliers?}\label{sec:pre_training_bert_medium}

We have now identified a correlation between the magnitude of hidden state parameters corresponding to outlier dimensions, and the frequency of the encoded tokens in the pre-training data. However, it is unclear whether the relationship is causal. 

To establish that, we pre-train\footnote{Except for the tokenization strategy, the training for each model is similar to the original BERT \cite{devlin2019bert} with two exceptions: (a) the Wikipedia corpus is a more recent version, from 01/03/2022, (b) the max sequence length is 256 (instead of 512) and batch size 128 instead of 256 due to computational constraints (these appear to have limited effect on the MNLI benchmark). All models were trained for 327,500 steps.} from scratch 3 versions of BERT-medium as defined by \citet{well-read-students}, with the following tokenization schemes:

\begin{itemize*}
    \item \split{}: We split sentences using a Spacy sentencizer\footnote{\url{https://spacy.io/}}, and add a `[SEP]' token at the end of each sentence and at the end of each encoded sequence. This is similar to the ``full sentences'' tokenization used to train RoBERTa \cite{liu2019roberta}.
    
    \item \onesep{}:
    We add a single `[SEP]' token at the end of each sequence of 256 tokens rather than each sentence. The main effect that we expect from this is that the amount of `[SEP]' tokens is reduced roughly by a factor of 10.
    
    \item \randomize{}: 
    Each sentence is followed by the `[SEP]' token, but we replace 50\% of occurrences of regular tokens with frequency above 1.e-5 in the training corpus with a random token with a frequency below 1.e-5 in 50\% of their occurrences\footnote{Due to high computational costs of BERT pre-training we only experiment with one possible value of the threshold (1.e-5). When exactly tokens become ``high frequency'' for BERT-type MLMs remains a question for future work.}.
\end{itemize*}

Note that the RoBERTa-like \split{} tokenization is different from the classic BERT approach, where each encoded sequence always contains exactly 2 `[SEP]' tokens in each encoded sequence. The RoBERTa approach would make this token more frequent for the sequences containing more than one sentence, and hence also more appropriate for testing our frequency hypothesis. %

Both \onesep{} and \randomize{} conditions corrupt the linguistic structure of the input, and so the trained MLM quality could be expected to drop as it acquires worse knowledge. But this setting will let us (a) identify the impact of token frequency on the outlier phenomenon, (b) disentangle the effect between frequent tokens in general and the `[SEP]' token.

All three models started from the same initialization but were fed different data according to the tokenization schemes. We find that \split{} model developed outliers \out{281} and \out{378}, whereas in \onesep{} the detrimental effect is only clear for \out{378}. The \randomize{} model developed two outliers: \out{353} and \out{362}.

\begin{table}[t]
    \centering
    \begin{adjustbox}{width=\linewidth}
    \input{tables/bert_medium_table}

    \end{adjustbox}
    \caption{Accuracy on MNLI-matched for each pre-training setting of BERT-medium.}
    \label{tab:bert_medium_table}
\end{table}

\Cref{tab:bert_medium_table} shows all three BERT-medium models evaluated on MNLI-matched validation set as either the full model or with their respective outliers removed one by one. \split{} model is the best performing overall, but \onesep{} is only .4 points behind as the full model. Both of them develop a very damaging outlier \out{378}, whereas the effect of \out{281} is less pronounced in \split{} and insignificant in \onesep{}. Moreover, the single outlier in \onesep{} is more damaging for the model. One possible explanation is that when the model has only one outlier, it likely relies on it more, which would result in higher performance degradation when it is disabled.

As expected, the \randomize{} model that was fed the noisiest data performs worse than the other two (by $\approx3\%$). But interestingly, it also does not develop any outliers as damaging as \out{378} is for the other two models.

We conclude that the frequency distribution of tokens in pre-training data contributes to the outlier phenomenon, and the `[SEP]' token is a part of that effect (since high frequency is one of the factors that characterizes it). 

%% file: figures/bert-base-uncased/broken_generated_frequencies.tex
\begin{figure}[t]
    \centering
    \subcaptionbox{\label{fig:bert_generated_tokens_broken} Without Outliers}[0.43\linewidth] {
        \includegraphics[width = \linewidth,height=0.25\textwidth]{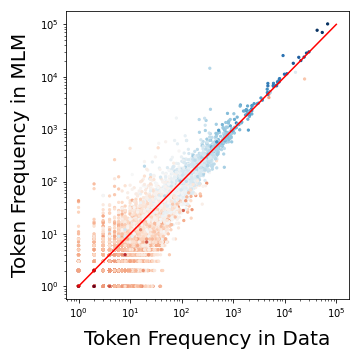}
    }
    \subcaptionbox{\label{fig:bert_generated_tokens_full} Full Model}[0.55\linewidth] {
        \includegraphics[width = \linewidth,height=0.25\textwidth]{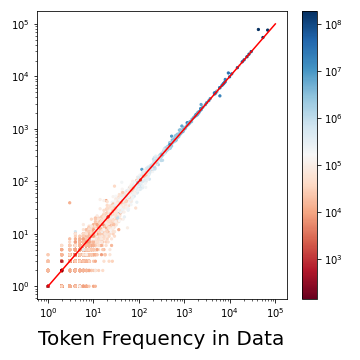}
    }
    \caption{
        A log-log scatter plot of token generation frequency vs true token frequency in data in MLM. The x-axis represents the number of time a token has been masked and the y-axis the times it has been predicted. The color shows the token appearances in pre-training data. In (\subref{fig:bert_generated_tokens_broken}) for the \emph{bert-base-uncased} model with zeroed out outliers and in (\subref{fig:bert_generated_tokens_full}) for the full pre-trained model. 
    }
    \label{fig:bert_generated_tokens}
\end{figure}

%% file: figures/bert-base-uncased/pre_trained_bert-base-uncased_correlation_with_frequency.tex
\begin{figure}[!t]
    \centering
    \subcaptionbox{\label{fig:pre_trained_outlier_corrtofreq_bert-base-uncased_w_special_tok} With special tokens}[0.49\linewidth]{
        \includegraphics[width = \linewidth,keepaspectratio]{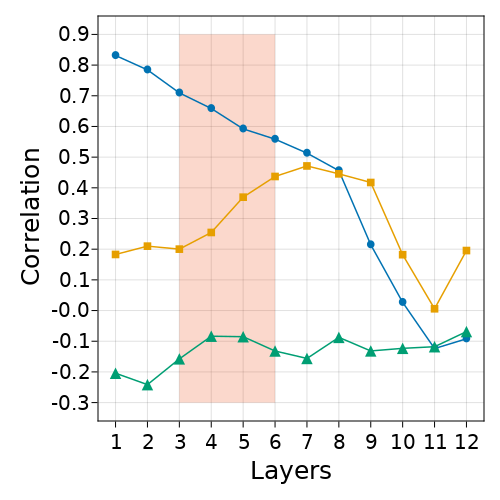}
    }
    \subcaptionbox{\label{fig:pre_trained_outlier_corrtofreq_bert-base-uncased_wo_special_tok} Without special tokens}[0.49\linewidth]{
        \includegraphics[width = \linewidth,keepaspectratio]{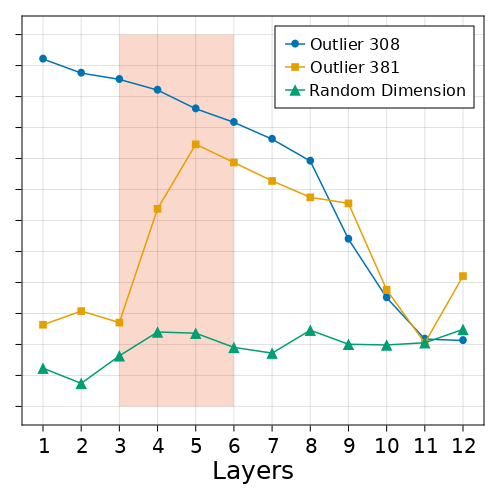}
    }
    \caption{
        BERT-base encoding Wikitext-v2 validation set: Pearson correlation between magnitude of hidden state parameters corresponding to outlier dimensions, and frequency of encoded tokens in pre-training data.  %
    }
    \label{fig:pre_trained_outlier_corrtofreq_bert-base-uncased}
\end{figure}

%% file: figures/bert-base-uncased/bert-base-uncased_nli_compared_to_mlm.tex
\begin{figure}[!t]
    \centering
    \subcaptionbox{\label{fig:bert-base-uncased_compared_nli} MNLI-m performance}[0.49\linewidth]{
        \includegraphics[width = \linewidth,keepaspectratio]{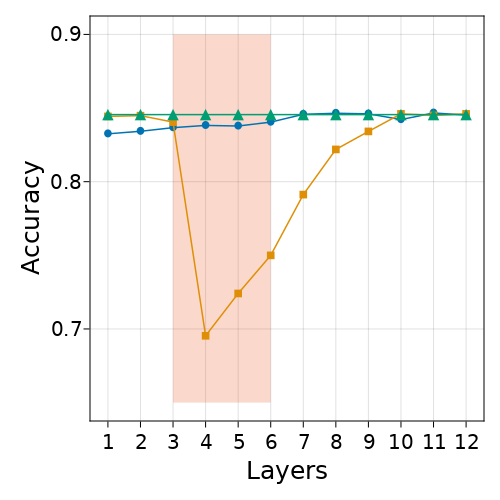}
    }
    \subcaptionbox{\label{fig:bert-base-uncased_compared_mlm} MLM loss (in wikitext-v2)}[0.49\linewidth]{
        \includegraphics[width = \linewidth,keepaspectratio]{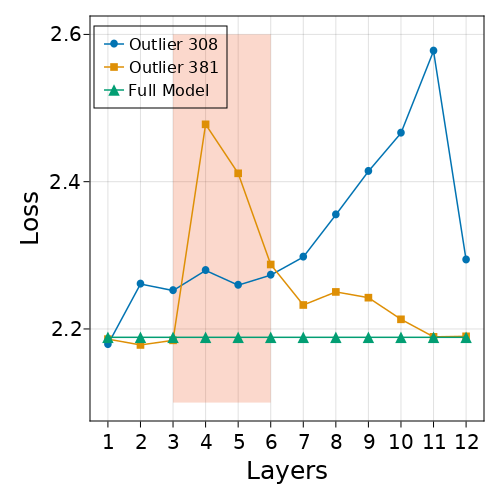}
    }
    \caption{BERT-base: effect of disabling outliers on MNLI-matched and MLM loss. 
    }
    \label{fig:bert-base-uncased_nli_compared_to_mlm}
\end{figure}

%% file: figures/bert-base-uncased/bert-base-uncased_attention_sample.tex
\begin{figure*}[!t]
    \captionsetup[subfigure]{labelformat=empty}
    \centering
    \begin{subfigure}[b]{\linewidth}
        \includegraphics[width = \linewidth,keepaspectratio]{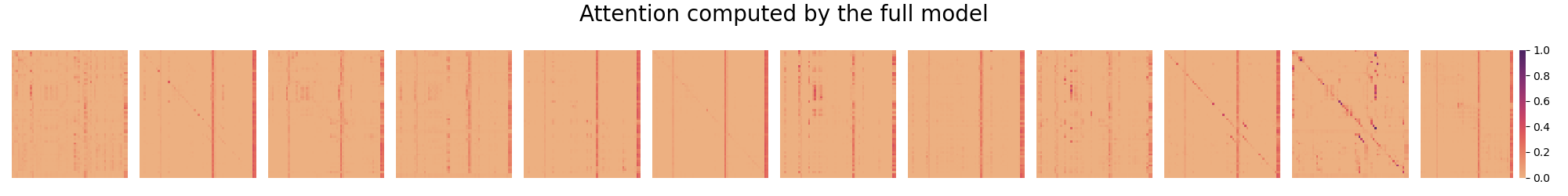}
    \end{subfigure}
    \begin{subfigure}[b]{\linewidth}
        \includegraphics[width = \linewidth,keepaspectratio]{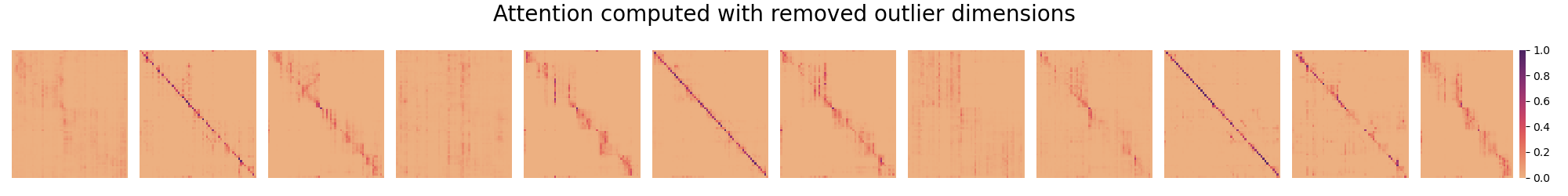}
    \end{subfigure}
    \caption{
        The self-attention patterns at the 10th layer of the full `bert-base-uncased' pre-trained model vs the same model with removed LayerNorm outliers. %
        \newline
Encoded example from MNLI: \emph{[CLS] Thebes held onto power until the 12th Dynasty, when its first king, Amenemhet I who reigned between 1980 1951 b.c. established a capital near Memphis.[SEP] The capital near Memphis lasted only half a century before its inhabitants abandoned it for the next capital. [SEP]}    
    }
    \label{fig:bert_qualitative_att_plot}
\end{figure*}

%% file: figures/bert-base-uncased/pre_trained_bert-base-uncased_correlation_with_attention.tex
\begin{figure*}[!t]
    \centering
    \subcaptionbox{\label{fig:pre_trained_bert_corr_with_att_308_with_spec_toks}\out{308}}[0.32\linewidth]{
        \includegraphics[width=\linewidth,keepaspectratio]{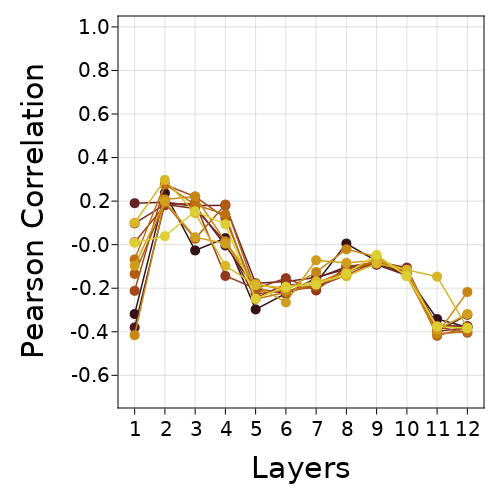}
    }
    \subcaptionbox{\label{fig:pre_trained_bert_corr_with_att_381_with_spec_toks} \out{381}}[0.32\linewidth]{
        \figuretitle{Including Special Tokens}
        \includegraphics[width=\linewidth,keepaspectratio]{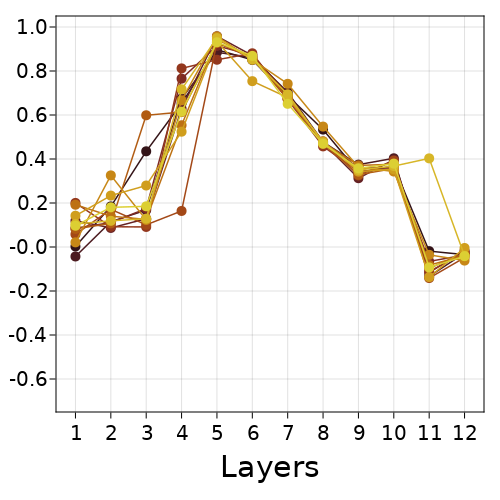}
    }
    \subcaptionbox{\label{fig:pre_trained_bert_corr_with_att_random_with_spec_toks} Random coefficients}[0.32\linewidth]{
        \includegraphics[width = \linewidth,keepaspectratio]{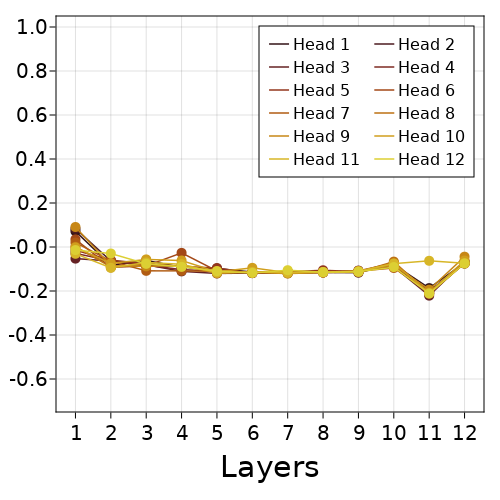}
    }
    \subcaptionbox{\label{fig:pre_trained_bert_corr_with_att_308_without_spec_toks} \out{308}}[0.32\linewidth]{
        \includegraphics[width=\linewidth,keepaspectratio]{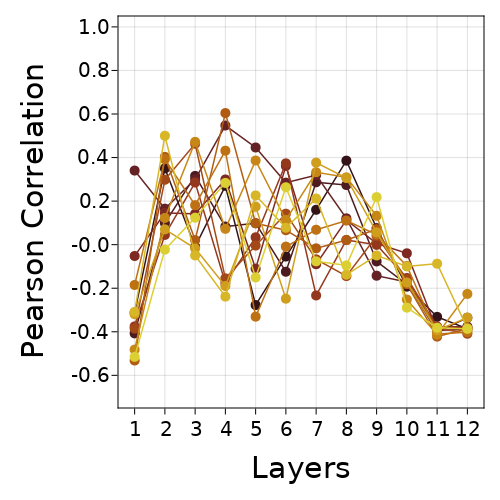}
    }
    \subcaptionbox{\label{fig:pre_trained_bert_corr_with_att_381_without_spec_toks} \out{381}}[0.32\linewidth]{
        \vspace{7pt}    
        \figuretitle{Excluding Special Tokens}
        \includegraphics[width = \linewidth,keepaspectratio]{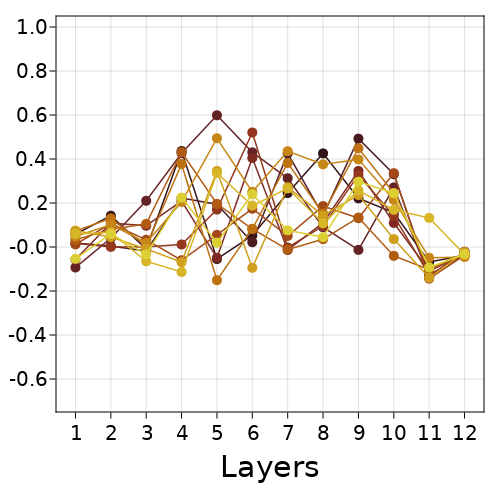}
    }
    \subcaptionbox{\label{fig:pre_trained_bert_corr_with_att_random_without_spec_toks} Random coefficients}[0.32\linewidth]{
        \includegraphics[width = \linewidth,keepaspectratio]{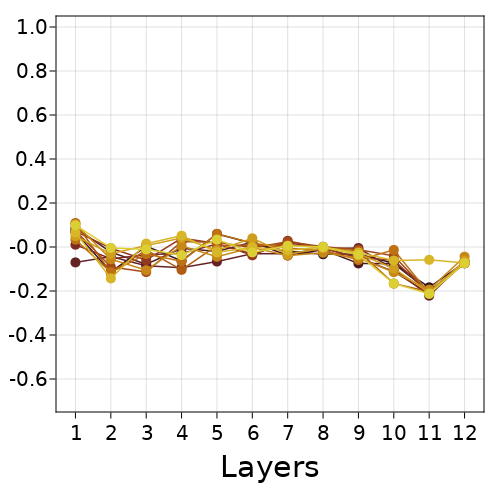}
    }
    
    \caption{Each figure shows the correlation between the \emph{average vertical attention values} in BERT-base self-attention heads, and the magnitude of  hidden state parameters at the dimensions corresponding to outlier dimensions. The correlation is computed over examples from Wikitext-v2. Fig. (\subref{fig:pre_trained_bert_corr_with_att_random_with_spec_toks}) and (\subref{fig:pre_trained_bert_corr_with_att_random_without_spec_toks}) show the average over 10 random dimensions.}
    \label{fig:pre_trained_bert_corr_with_att}
\end{figure*}

%% file: tables/bert_medium_table.tex
\begin{tabular}{cccc}
\toprule
{} & \split{} & \onesep{} & \randomize{}\\
\midrule
Full Model & 79.6 & 79.2 & 76.8 \\
\midrule
Minus \out{378} & \textbf{66.9} & \textbf{47.4} & -\\
Minus \out{281} & 78.8 & - & -\\
Minus \out{353} & - & - & 75.8\\
Minus \out{362} & - & - & \textbf{74.4}\\
\bottomrule
\end{tabular}

%% file: chapters/5_discussion.tex
\section{Discussion}

\subsection{Outliers in Transformer Pre-training}

Prior work \cite{KovalevaKulshreshthaEtAl_2021_BERT_Busters_Outlier_Dimensions_that_Disrupt_Transformers} showed that the outliers are present in a large number of Transformer-based LMs. We provide complementary evidence for the Vision Transformer (\cref{tab:vit_outliers}). However, we were unable to identify outliers in protein and audio Transformers, which we attribute to significantly smaller vocabulary size. This finding hints towards the training data distribution being at the core of the outlier phenomenon.

\citet{KovalevaKulshreshthaEtAl_2021_BERT_Busters_Outlier_Dimensions_that_Disrupt_Transformers} also show that outliers emerge early in pre-training (after 50K steps for BERT-medium). 
We extend that experiment by investigating the fully pre-trained BERT-base by \citet{sellam2022the-multiberts}, we find that the impact of outliers on the model grows up steadily until step $10^6$. After that step the outlier effect is inconsistent between checkpoints, and the full model performance saturates. An interesting question for future work is what happens after outlier removal stops degrading model performance (around step $10^6$), and whether it could be used as an early stopping criterion.

Transformer-based language models (LMs) have been shown to exhibit anisotropic behavior in their representations of both tokens and sentences \cite{,Ethayarajh_2019_How_Contextual_are_Contextualized_Word_Representations_Comparing_Geometry_of_BERT_ELMo_and_GPT-2_Embeddings,gaoetal-representation_degeneration, RajaeePilehvar_2021_How_Does_Fine-tuning_Affect_Geometry_of_Embedding_Space_Case_Study_on_Isotropy,TimkeyvanSchijndel_2021_All_Bark_and_No_Bite_Rogue_Dimensions_in_Transformer_Language_Models_Obscure_Representational_Quality}. While pervasive, this is an undesirable property because it reduces the average distance between tokens embeddings, and thus makes it more difficult to distinguish between tokens in the embedding space.

One consequence of the outliers growth over pre-training is that the attempts to remove anisotropy at the downstream task level 
\cite{RajaeePilehvar_2021_How_Does_Fine-tuning_Affect_Geometry_of_Embedding_Space_Case_Study_on_Isotropy}, although effective in some cases, could be only partially addressing the problem. In that case it might be more productive to change pre-training so as to better account for the skewed token frequency distribution \cite{frequency_aware_sgd}.

\subsection{Outliers and Token Frequency}

\citet{li-etal-2021-bert} and \citet{gaoetal-representation_degeneration} show that embeddings of low frequency tokens lie further away from high frequency ones in the embedding space. 
In \cref{sec:influence_of_frequency} we showed how the outlier parameters influence the hidden state geometry proportionally to token frequency, and how this is more sensitive at earlier layers. This is consistent with findings of \citet{li-etal-2021-bert} who show that the different geometry of frequent and non-frequent tokens is more evident for the layers closer to the input. Indeed, we observe this effect for \out{381} in BERT-base and \out{588} in RoBERTa-base (see \cref{fig:pre_trained_outlier_corrtofreq_bert-base-uncased}).

To the best of our knowledge, this is the first work to demonstrate the link not only between the geometry of the hidden states and frequency of encoded tokens in pre-training data, but also the model performance.

\subsection{Outliers and Positional Embeddings}

Concurrently with the demonstration of the outlier phenomenon by \citet{KovalevaKulshreshthaEtAl_2021_BERT_Busters_Outlier_Dimensions_that_Disrupt_Transformers}, \citet{LuoKulmizevEtAl_2021_Positional_Artefacts_Propagate_Through_Masked_Language_Model_Embeddings} attributed the high-magnitude weights to a different source: positional embeddings rather than LayerNorm weights. The positional embeddings could be expected to have more impact in the earlier layers. Our work contributes to the dispute by showing that two different behaviours are present in both BERT-base and RoBERTa-base: one outlier dimension in the hidden states is disruptive in layers 4-6 (\out{381} for BERT and \out{588} for RoBERTa) while the other one at the layers 10-11 (\out{308} for BERT and \out{77} for RoBERTa). %
This suggests that both mechanisms may play a role.

\subsection{Outliers and Self-Attention}

\citet{kovaleva-etal-2019-revealing} identify 5 frequent self-attention patterns, 4
of which include vertical lines corresponding to special tokens. We showed (\cref{fig:pre_trained_bert_corr_with_att}) that the presence of special tokens increases the correlation (in absolute value) between the average query value and the magnitude of the hidden state dimensions corresponding to outlier dimensions. This suggests that the outlier phenomenon contributes to the vertical attention patterns identified by \citet{kovaleva-etal-2019-revealing}. 
From the computational perspective this is consistent with the attention being a bilinear form. Moreover, the relation between the outliers and the vertical self-attention pattern (often ``pointing'' to the highly frequent special tokens and punctuation) also hints at the relation between outliers and the token distribution in the pre-training data.

At the same time, the correlation remains evident in the final layers even when special tokens are ignored, indicating that the outliers also contribute to the attention shape more broadly. This is in line with \citet{kobayashi-etal-2020-attention} who argue that vertical patterns in attention do not indicate that no other information is encoded (hence simply norming the self-attention makes other relations more salient).

%% file: chapters/6_conclusion.tex
\section{Conclusion}

To the best of our knowledge, this is the first work to directly link the outlier dimension phenomenon in Transformer-based models (in particular BERT and RoBERTa) to encoded token frequency in pre-training data. We also find that the magnitude of hidden state dimensions corresponding to outliers correlates with the vertical self-attention pattern, which enables the attention to the classification tokens. Furthermore, we find that there are two types of outliers: some of them affect the MLM performance the most in the middle layers (where the correlation with token frequency is also at its peak), and for others the impact grows towards the final layers (even though the correlation with token frequency decreases).

Our findings suggest that outliers are due not to the Transformer architecture per se, but rather to the highly skewed token frequency distribution in textual pre-training data. In that case, to mitigate anisotropy we might need to design a pre-training scheme that better accounts for such distributions.

%% file: chapters/7_limitations.tex
\section{Limitations}

This work establishes a relation between the outlier phenomenon in Transformer-based language models and the frequency of tokens in the corpus used for pre-training. We focus on two of the most popular Transformers (BERT-base and RoBERTa-base) and show that our key observations hold for both of them, but there are hundreds of other possible Transformer-based LM architectures and modifications to pre-training regimes and tokenization that could be explored in future work. That being said, we believe that a thorough examination of the most common models, such as presented in this paper, is pre-requisite to establishing the methods and hypotheses for a large-scale study.

Methodologically, our experiments have the following limitations:

\begin{itemize}
    \item We identify a correlation between frequency of the encoded tokens in the pre-training corpus, and the magnitude of the coefficients within the hidden states of Transformer layers which correspond to the outlier dimensions (\cref{sec:influence_of_frequency}). A limitation of this experiment is that it only establishes that there is mutual influence between token frequency and outliers. We cannot exclude the presence of covariates or claim that the token distribution is the reason behind the outlier phenomenon.
    \item To establish whether there is also a causal relation, we show that pre-training a \emph{bert-medium} model with changes in the tokenization reduces the impact of outlier removal (\cref{sec:pre_training_bert_medium}). This experiment does show that we can almost remove the outliers effect by changing the token distribution, however, these changes by themselves degrade the quality of the model and its downstream task performance, and therefore the reduced outlier removal effect could be partly due to the model being overall less performant. This also raises a follow-up question for future work: do the special tokens have such a connection to the outlier phenomenon partly because of their special role, or simply due to the fact that they are among the most frequent tokens in the pre-training corpus?
    \item We find that outliers have a strong impact on the shape of attention heads, most notably the ``vertical'' patterns (\cref{sec:attention}), and we hypothesize that the outlier removal effect on downstream task performance may thus be explained by the inability of the model to ``focus'' on the special tokens, which according to prior work is a key role of the ``vertical'' self-attention patterns. This hypothesis merits further investigation.
\end{itemize}

%% file: chapters/8_broader_impact.tex
\section{Broader Impacts}

This work focuses on the analysis of two popular Transformer-based LMs of the BERT family (BERT- and RoBERTa-base). This work relies on established benchmarks, does not collect new human subjects data and presents no new models. Its broader impacts center on improving the mechanistic understanding of training Transformer-based LMs, which could lead to developing better models in the future. We also presented evidence of outlier phenomenon in Vision Transformer, which suggests that vision and multimodal Transformers may also be vulnerable to attacks involving direct modification of outlier weights.

Experiments were conducted using a private infrastructure, which has a estimated carbon efficiency of 0.37 kgCO$_2$eq/kWh (average carbon efficiency in Japan, where the machine is based, for the year 2020). Including experiments that were discarded and failed runs, we estimate that a cumulative of 200 hours of computation was performed on hardware of type RTX A6000 (TDP of 300W). Total emissions are estimated to be 22.2 kgCO$_2$eq.

%% file: chapters/appendix.tex
\appendix

\section{Outliers For Each Model}
\label{app:outlier-list}

\begin{table}[!t]
    \centering
    \begin{adjustbox}{width=\linewidth}
    \input{tables/outliers_by_model}
    \end{adjustbox}
    \caption{The outliers identified for each model used in the paper}
    \label{tab:outliers_by_model}
\end{table}

As described in \cref{sec:outlier-def} the outliers are zeroed out in the LayerNorm layers,
in \cref{sec:outlier-def} we mention that the definition of outliers is not entirely formal: while the weights magnitude let us identify a small subset of weight among which we can search for outliers, we need to fine tune the model on a downstream task, we use MNLI although all tasks in GLUE would work, to identify which weights are the most harmful. %
\cref{tab:outliers_by_model} lists the two most damaging outliers for each model we used in the paper.

\section{Replicability}
\label{app:reproducibility}

In this section we describe in detail all the experiments carried out within this work, together with the code used to make them this should allow for an effective reproduction of our results. All experiments are carried out using a NVidia A6000 with 48 gbyte of memory.

\begin{itemize}
    \item For the results in \cref{tab:bert-outliers} we fine tune \emph{bert-base-uncased} for 4 epochs on each task with a 2.e-5 learning rate and 256 maximum sequence length. We measure the respective metric for each GLUE task (as defined by \citet{wang-etal-2018-glue}) on the validation set. Both models and datasets are loaded through huggingface \url{https://huggingface.co/}. For the computation with removed outliers, what we do is we compute the same metric as for the full model after manually setting to $0$ the chosen LayerNorm \emph{weight} and \emph{bias} parameters in all layers. A similar procedure is adopted to compute the values in \cref{tab:vit_outliers}. Fine tuning on the largest datasets within the glue benchmarks (mnli, qnli, qqp),  with the hyperparameters described above on average requires approximately 4000 seconds. The remaining datasets among the glue benchmarks are between 10 to 100 times smaller and require a proportionally scaled amount of time.

    \item The token counts in \cref{fig:roberta_generated_tokens} are obtained through Wikipedia and book corpus by directly using a \emph{bert-base-uncased} and \emph{roberta-base} tokenizers on the whole corpus and counting each token occurrence.

    \item The results in \cref{fig:pre_trained_outlier_corrtofreq_bert-base-uncased} are obtained as follows: for each token in the data (as part of an encoded sequence) we compute the hidden states through a \emph{bert-base-ucased} model and pick the hidden state parameter at the outlier index therefore getting a single numerical value for each token. We also associate to each token its frequency in the pre-training corpus and we measure the Pearson correlation coefficient between this two lists of values.

    \item The results in \ref{fig:bert-base-uncased_nli_compared_to_mlm} are obtained by setting LayerNorm \emph{weight} and \emph{bias} parameters at the given outlier index for a given layer to $0$. For \cref{fig:bert-base-uncased_compared_nli} this is done for a model fine-tuned on MNLI train set and we measure the accuracy on MNLI matched, for \cref{fig:bert-base-uncased_compared_mlm} this is done to a pre-trained only model by measuring the MLM loss on the wikitext-v2 validation set (the masking probabilities are kept as in the original BERT paper). This process is repeated for each layer in the model.

    \item The results in \cref{fig:pre_trained_bert_corr_with_att} are obtained as follows: for each sample in the wikitext-v2 validation set (a single sequence containing $n$ tokens), we encode it with \emph{bert-base-uncased}. This provides us with attention matrices with size $n \times n$ we take the average over the columns, thus getting a single numerical value for each token. As above, for each token we also collect the hidden state value at the outlier dimension, a single numerical value (for each layer) for each token, and finally we measure the correlation between these two values. In particular since for each layer there are 12 heads we compute 12 correlations at each layer.

    \item The scores in \cref{tab:bert_medium_table} are obtained as for \cref{tab:bert-outliers} on three instances of \emph{bert-medium} architecture pre-trained with different tokenization strategies. The pretraining of this model is performed with 256 max length, 128 batch size and 1.e-4 learning rate.

\end{itemize}

\input{figures/roberta-base/broken_generated_frequencies}

Experiments were conducted using a private infrastructure, which has a estimated carbon efficiency of 0.37 kgCO$_2$eq/kWh (average carbon efficiency in Japan, where the machine is based, for the year 2020). Including experiments that were discarded and failed runs, we estimate that a cumulative of 200 hours of computation was performed on hardware of type RTX A6000 (TDP of 300W). Total emissions are estimated to be 22.2 kgCO$_2$eq.

\begin{table}[!t]
    \centering
    \begin{adjustbox}{width=\linewidth}
    \input{tables/roberta_base_full}
    \end{adjustbox}
    \caption{Full RoBERTa scores on GLUE benchmarks with outlier effects.}
    \label{tab:roberta-outliers-full-glue}
\end{table}

\section{RoBERTa Experiments}
\label{app:roberta}

The results we showed for BERT-base similarly hold for RoBERTa-base. The generation distribution with removed outliers, \cref{fig:roberta_generated_tokens_broken}, shows that a single token on the top right, the "</s>" token, is generated a larger number of times (log scale), making this coherent with the results for BERT. We note that RoBERTa pre-training data is not the same as we use\footnote{For RoBERTa the token frequency in pre-training is computed on Wikipedia + Book corpus plus an open source version of OpenWebText (\url{https://huggingface.co/datasets/openwebtext}), however RoBERTa pre-training data also include Stories and CC-news datasets not openly available.}, however the core of the vocabulary is shared and therefore the qualitative results shown in \cref{fig:roberta_generated_tokens} are reliable.

\cref{tab:roberta-outliers-full-glue} shows the performance degradation with outliers removed on all GLUE tasks. 
As shown by \citet{KovalevaKulshreshthaEtAl_2021_BERT_Busters_Outlier_Dimensions_that_Disrupt_Transformers} there is one more damaging outlier \out{588} and a less damaging one \out{77}, when removed together they cause the largest performance degradation. \cref{fig:pre_trained_outlier_corrtofreq_roberta_base_w_special_tok} and \cref{fig:roberta-base_nli_compared_to_mlm} show that for RoBERTa patterns similar to those we see for BERT in \cref{fig:pre_trained_outlier_corrtofreq_bert-base-uncased} and \cref{fig:bert-base-uncased_nli_compared_to_mlm} appear. 

In particular, \out{588} is more damaging when the magnitude of the respective hidden state outlier dimension correlates the most to token frequency. In this case at layers 2-4 and at layer 10. In \cref{fig:pre_trained_outlier_corrtofreq_roberta_base_wo_special_tok} we observe a spike in correlation with frequency, and \cref{fig:roberta-base_compared_mlm} shows a similar one for MLM loss. On the other hand, \out{77} shows that the less the hidden state dimension corresponding to the outlier correlates to frequency, the more the removal of the LayerNorm outlier damages the model. 

\input{figures/roberta-base/pre_trained_roberta-base_correlation_with_frequency.tex}
\input{figures/roberta-base/roberta-base_nli_compared_to_mlm.tex}

\input{figures/roberta-base/pre_trained_roberta-base_correlation_with_attention}

For this model we also see an anti-pattern at layer 4 (\cref{fig:roberta-base_compared_mlm}): the loss with \out{77} is higher and the one with \out{588} is lower. However, \cref{fig:pre_trained_outlier_corrtofreq_roberta_base_w_special_tok} shows that layer 4 is where the correlation including special tokens is closest to zero, possibly due to RoBERTa pre-training schedule including a larger number of special tokens \cite{liu2019roberta}.

The general pattern observed for BERT is kept, however, while for BERT the worst layers in term of performance are layers 4-5, for RoBERTa this are layers closer to the input 1-2. One of the reasons behind this difference could be that RoBERTa had longer pre-training.

Finally we also replicate the analysis of attention patterns. \cref{fig:pre_trained_rob_corr_with_att} shows for RoBERTa the same patters that \cref{fig:pre_trained_bert_corr_with_att} shows for BERT: for the hidden state parameters corresponding to the outlier dimensions, the correlation values are very different when compared to random ones, both when including the special tokens or not.

\section{Outliers in Pre-Training}
\label{app:pre-training}

\input{figures/multiberts_over_pretraining/multiberts_mnli_mismatched_over_pretraining}

\Cref{fig:multiberts_seed_1_mnli_mm_acc_alone}  shows the accuracy on MNLI-mismatched, at various checkpoints for BERT-base seed 1 provided \citet{sellam2022the-multiberts}. The results are very similar to what \cref{fig:multiberts_seed_1_mnli_acc_alone} shows for MNLI-matched: early degradation around 80,000 steps, almost steadily worsening until step 1,000,000, and then fluctuating further on. The initialization of the classification layer is not fixed across checkpoints. 

\input{figures/multiberts_over_pretraining/fixed_seed_mnli_over_pretraining}

In \cref{fig:fixed_seed_multiberts_seed_1_mnli_acc_alone} and \cref{fig:fixed_seed_multiberts_seed_1_mnli_mm_acc_alone} we also replicate the experiments while fixing the classification head seed at initialization. In this case as well the results are very close to those from \cref{fig:multiberts_seed_1_mnli_acc_alone} and \cref{fig:multiberts_seed_1_mnli_mm_acc_alone}. Specifically, the fluctuating behaviour appearing after 1 million steps is very distinct in this case as well. It is therefore not caused by changes in different fine-tuning initialization but confirmed to be caused by the number of pre-training steps. 

An interesting question for future research on this topic is, what is the influence of longer pre-training on this phenomenon, does it get slowly cancelled? Does adding pre-training data from sources other than Wikipedia, the largest source of data for the models we investigate, make the outliers effect smaller or larger?

\section{POS Tag Distribution of Tokens Predicted by BERT MLM with Disabled Outliers}\label{app:outliers-pos-shift}

In this experiment we investigated the POS tags of the tokens predicted by the BERT-base MLM with disabled outlier dimensions. \cref{fig:bert-base-uncased_outlier_pos_shift} shows the distribution of tags over the replaced tokens. Each row shows the percentage of tags of generated tokens with respect to the tag of the masked token: for example, the top row in \autoref{fig:bert-base-uncased_outlier_pos_shift_308_381} shows that ADJ tokens are replaced with 16\% probability by NOUN tokens, with 35\% by ADJ tokens, with 10\% by PUNCT tokens and so on.

\input{figures/multiberts_over_pretraining/fixed_seed_mnli-mm_over_ptretraining}

We have previously shown in \autoref{tab:bert-outliers} that out of two outliers one damages the model performance considerably more. This pattern is also observed here. \cref{fig:bert-base-uncased_outlier_pos_shift} shows that individually \out{381} has a much larger effect than \out{308}. We can also see the qualitative difference between the outliers in the distribution of POS tags of the generated tokens: with only \out{381} disabled, the model becomes more likely to generate nouns and punctuation signs, while \out{308} does not produce so many changes. However, \out{308} has a larger effect in combination with \out{381}, again pushing the model towards generating more nouns and punctuation, but also symbols and adpositions.

\clearpage
\input{figures/bert-base-uncased/bert-base-uncased_pos_shift.tex}
\clearpage

\section{Outliers vs Encoded Token Frequency: the Case of Fine-Tuned Models}
\label{app:fine-tuning}

\input{figures/bert-base-uncased/fine_tuned_bert-base-uncased_correlation_with_frequency}

To control how much fine-tuning affects the patterns we study, we repeat the experiments with models fine-tuned on MNLI, we proceed as follows: we fine-tune the model using a classification head and then extract the hidden states at each layer and use those in place of the ones of the pre-trained model.

\Cref{fig:fine_tuned_outlier_corrtofreq_bert-base-uncased} shows the same information as \cref{fig:pre_trained_outlier_corrtofreq_bert-base-uncased}, that is the correlation between the hidden states outlier dimension magnitude and the frequency of the encoded tokens in pre-training data, for a BERT-base model fine-tuned on MNLI. The overall patterns is similar to using the pre-trained model, but the correlation values generally decrease: the highest value is now 0.3 when it used to be 0.5 for the pre-trained model. This agrees with the findings from \cref{sec:pre_training_bert_medium}: the outliers are impacted by the model training.

Investigating attention patterns, \cref{fig:fine_tuned_bert_corr_with_att} reports the same information as \cref{fig:pre_trained_bert_corr_with_att} for \emph{bert-base-uncased} model fine-tuned on MNLI. In this case we see that the correlations stays high at layers closer to the input data, while those closer to the output have lower values, although in this case as well the values are higher than they are for random outliers \cref{fig:fine_tuned_bert_corr_with_att_random_with_spec_toks,fig:fine_tuned_bert_corr_with_att_random_without_spec_toks}. 

\clearpage
\input{figures/bert-base-uncased/fine_tuned_bert-base-uncased_correlation_with_attention}

%% file: tables/outliers_by_model.tex
\begin{tabular}{lcc}
\toprule
\emph{Model name}      &  Outlier 1 & Outlier 2\\
\midrule
"bert-base-uncased"    &  308 & 381 \\
"roberta-base"         &   77 & 588 \\
"multiberts-seed-1"    &  218 & 674 \\
"google/vit-095base-patch16-224-in21k" & 187 & 759 \\
"BERT-medium (ours)" & 281 & 378\\
\bottomrule
\end{tabular}

%% file: figures/roberta-base/broken_generated_frequencies.tex
\begin{figure}[t]
    \centering
    \subcaptionbox{\label{fig:roberta_generated_tokens_broken} Without Outliers}[0.43\linewidth] {
        \includegraphics[width = \linewidth,height=0.24\textwidth ]{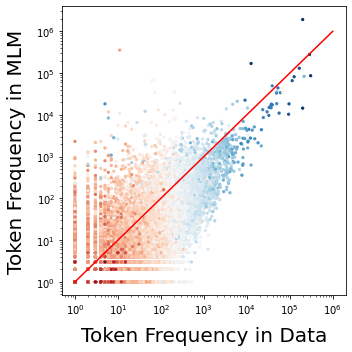}
    }
    \subcaptionbox{\label{fig:roberta_generated_tokens_full} Full Model}[0.55\linewidth] {
        \includegraphics[width = \linewidth, height=0.24\textwidth]{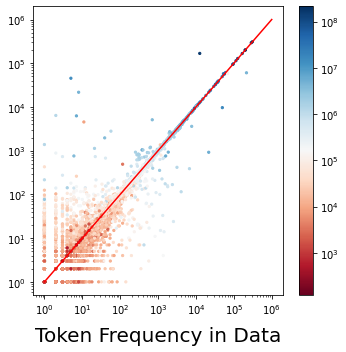}
    }
    \caption{
        A log-log scatter plot of token generation frequency vs true token frequency in data in MLM. The x-axis represents the number of time a token has been masked and the y-axis the times it has been predicted. The color shows the token frequency in pre-training data (wikipedia + book corpus). In (\subref{fig:roberta_generated_tokens_broken}) for the \emph{roberta-base} model with zeroed out outliers and in (\subref{fig:roberta_generated_tokens_full}) for the pre-trained model. 
    }
    \label{fig:roberta_generated_tokens}
\end{figure}

%% file: tables/roberta_base_full.tex
\begin{tabular}{crrrrrrrrr}
\toprule
\emph{Outliers} &  cola &  mnli-mm &  mnli &  mrpc &  qnli &   qqp &   rte &  sst2 &  stsb \\
\midrule
baseline    &  58.3 &     87.4 &  87.6 &  87.3 &  92.7 &  91.4 &  69.0 &  95.0 &  89.1 \\
\midrule
77      &  51.5 &     85.4 &  85.5 &  80.1 &  89.8 &  90.4 &  65.0 &  93.9 &  83.7 \\
588     &   7.4 &     61.5 &  59.4 &  70.8 &  56.6 &  64.2 &  54.2 &  70.3 &  19.1 \\
588, 77 &  12.6 &     45.9 &  44.9 &  70.3 &  50.6 &  61.2 &  51.6 &  68.8 &   5.4 \\
\bottomrule
\end{tabular}

%% file: figures/roberta-base/pre_trained_roberta-base_correlation_with_frequency.tex
\begin{figure}[!t]
    \centering
    \subcaptionbox{\label{fig:pre_trained_outlier_corrtofreq_roberta_base_w_special_tok} With special tokens}[0.49\linewidth]{
        \includegraphics[width = \linewidth,keepaspectratio]{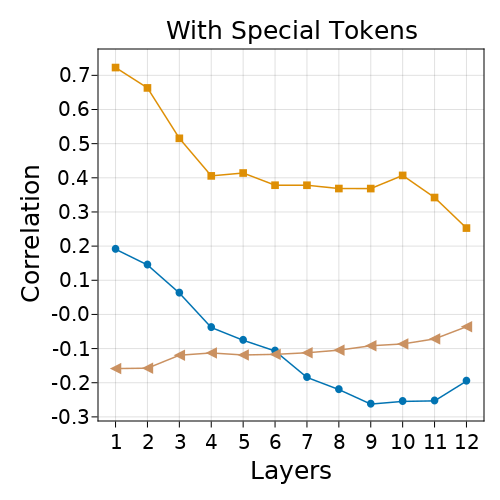}
    }
    \subcaptionbox{\label{fig:pre_trained_outlier_corrtofreq_roberta_base_wo_special_tok} Without special tokens}[0.49\linewidth]{
        \includegraphics[width = \linewidth,keepaspectratio]{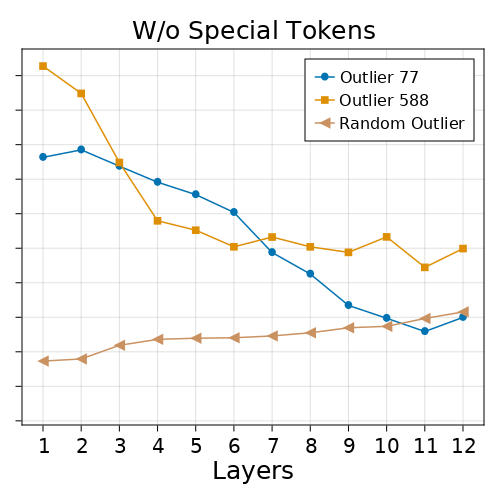}
    }
    \caption{
        Correlation of outlier dimension magnitude with token frequency over the Wikitext corpus for a pre-trained RoBERTa-base model. In (\subref{fig:pre_trained_outlier_corrtofreq_roberta_base_w_special_tok}) the correlations accounts for special tokens, in (\subref{fig:pre_trained_outlier_corrtofreq_roberta_base_wo_special_tok}) they are excluded.
    }
    \label{fig:pre_trained_outlier_corrtofreq_roberta_base}
\end{figure}

%% file: figures/roberta-base/roberta-base_nli_compared_to_mlm.tex
\begin{figure}[!t]
    \centering
    \subcaptionbox{\label{fig:roberta-base_compared_nli} MNLI-m performance}[0.49\linewidth]{
        \includegraphics[width = \linewidth,keepaspectratio]{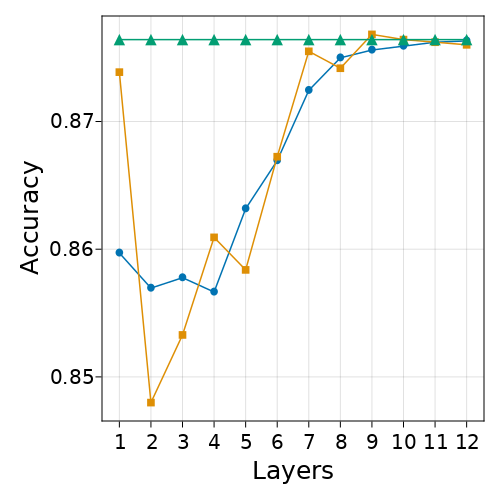}
    }
    \subcaptionbox{\label{fig:roberta-base_compared_mlm} MLM loss (in wikitext-v2)}[0.49\linewidth]{
        \includegraphics[width = \linewidth,keepaspectratio]{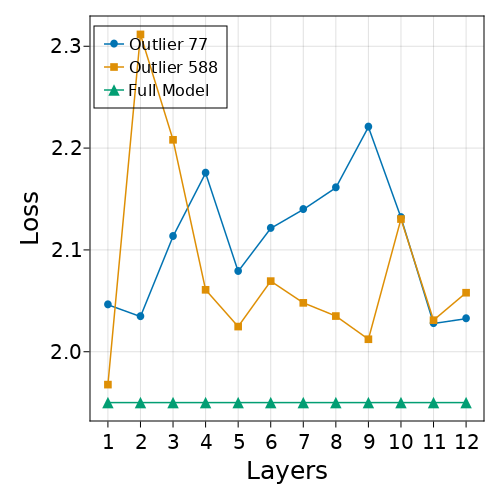}
    }
    \caption{RoBERTa-base: effect of disabling outliers.
    }
    \label{fig:roberta-base_nli_compared_to_mlm}
\end{figure}

%% file: figures/roberta-base/pre_trained_roberta-base_correlation_with_attention.tex
\begin{figure*}[!ht]
    \centering
    \subcaptionbox{\label{fig:pre_trained_rob_corr_with_att_77_with_spec_toks} \out{77}}[0.32\linewidth]{
        \includegraphics[width = \linewidth,keepaspectratio]{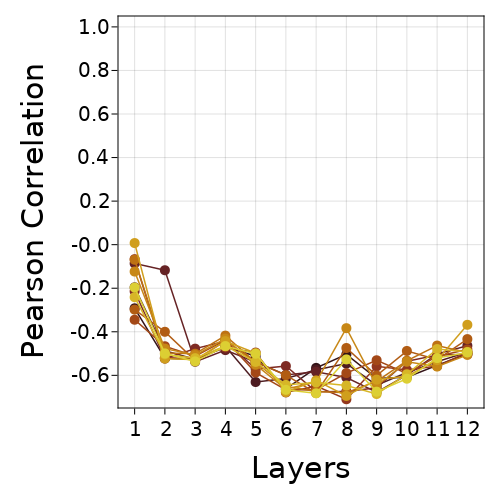}
    }
    \subcaptionbox{\label{fig:pre_trained_rob_corr_with_att_588_with_spec_toks} \out{588}}[0.32\linewidth]{
        \figuretitle{Including special tokens}
        \includegraphics[width = \linewidth,keepaspectratio]{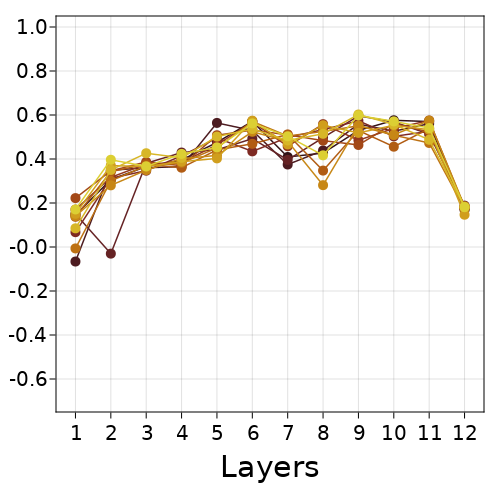}
    }
    \subcaptionbox{\label{fig:pre_trained_rob_corr_with_att_random_with_spec_toks} Random coefficients}[0.32\linewidth]{
        \includegraphics[width = \linewidth,keepaspectratio]{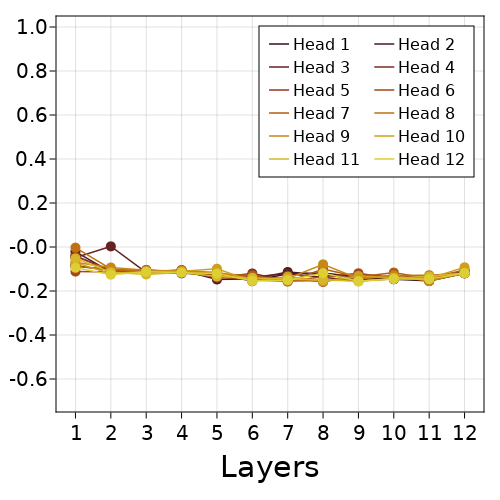}
    }
    \subcaptionbox{\label{fig:pre_trained_rob_corr_with_att_77_without_spec_toks} \out{77}}[0.32\linewidth]{
        \includegraphics[width = \linewidth,keepaspectratio]{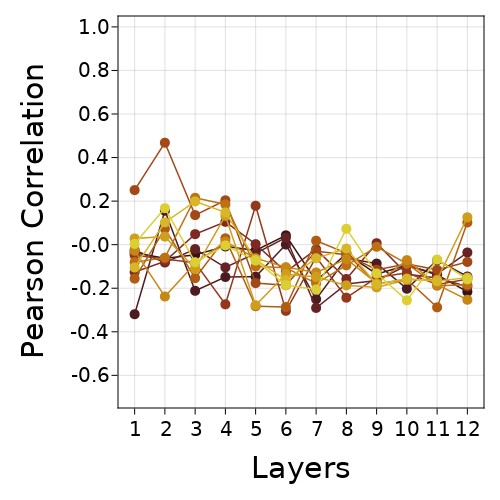}
    }
    \subcaptionbox{\label{fig:pre_trained_rob_corr_with_att_588_without_spec_toks} \out{588}}[0.32\linewidth]{
        \vspace{7pt}
        \figuretitle{Excluding special tokens}
        \includegraphics[width = \linewidth,keepaspectratio]{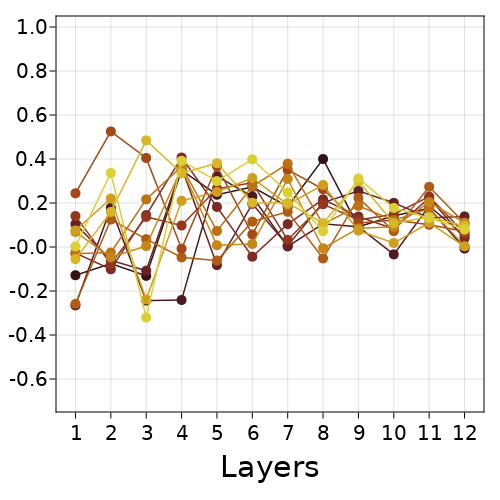}
    }
    \subcaptionbox{\label{fig:pre_trained_rob_corr_with_att_random_without_spec_toks} Random coefficients}[0.32\linewidth]{
        \includegraphics[width = \linewidth,keepaspectratio]{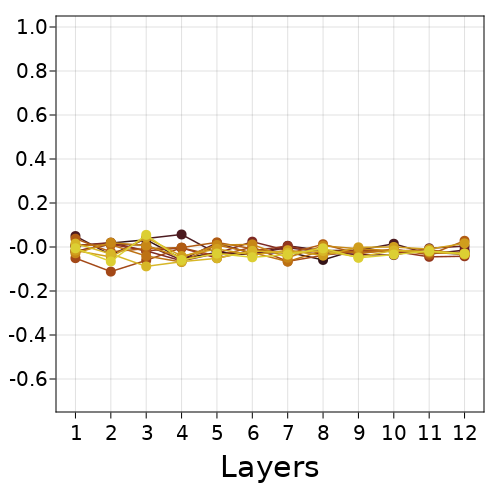}
    }
    
    \caption{Each figure shows the correlation between the \emph{average vertical attention values} in RoBERTa-base self-attention heads, and the magnitude of  hidden state parameters at the dimensions corresponding to outlier dimensions. The correlation is computed over examples from Wikitext-v2. Figures (\subref{fig:pre_trained_bert_corr_with_att_random_with_spec_toks}) and (\subref{fig:pre_trained_bert_corr_with_att_random_without_spec_toks}) show the average over 10 random dimensions.}
    \label{fig:pre_trained_rob_corr_with_att}
\end{figure*}

%% file: figures/multiberts_over_pretraining/multiberts_mnli_mismatched_over_pretraining.tex
\begin{figure}[!ht]
    \centering
    \includegraphics[width = \linewidth,keepaspectratio,trim=0 0 0 48,clip]{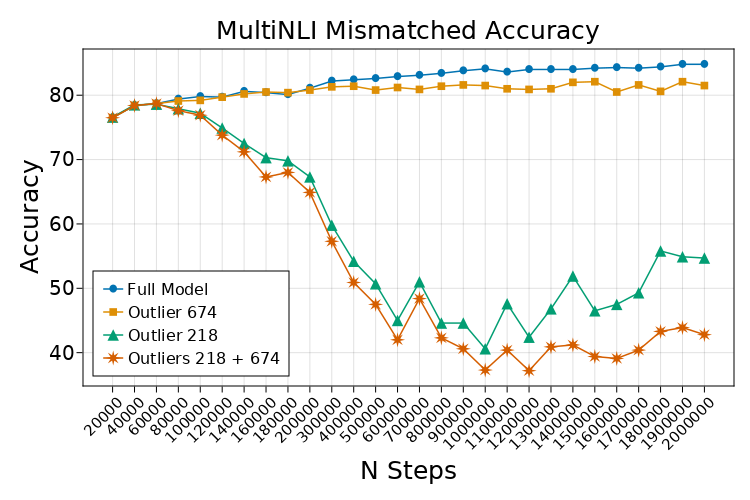}
    \caption{
        The accuracy on MNLI-mismatched of the checkpoints for BERT-base (seed 1), provided by \citet{sellam2022the-multiberts}.
    }
    \label{fig:multiberts_seed_1_mnli_mm_acc_alone}
\end{figure}

%% file: figures/multiberts_over_pretraining/fixed_seed_mnli_over_pretraining.tex
\begin{figure}[!t]
    \centering
    \includegraphics[width = \linewidth,keepaspectratio,trim=0 0 0 45,clip]{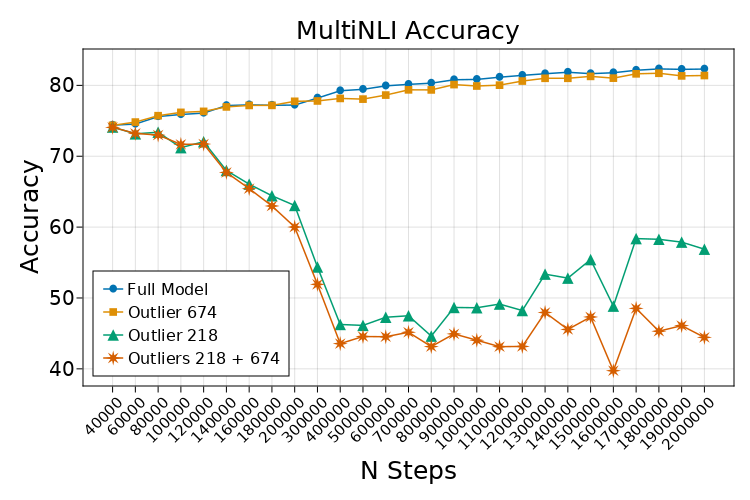}
    \caption{
        The accuracy on MNLI-matched of the checkpoints for BERT-base (seed 1) by \citet{sellam2022the-multiberts} for full model or with each outlier removed. All classification heads are equally initialized.
    }
    \label{fig:fixed_seed_multiberts_seed_1_mnli_acc_alone}
\end{figure}

%% file: figures/multiberts_over_pretraining/fixed_seed_mnli-mm_over_ptretraining.tex
\begin{figure}[!t]
    \centering
    \includegraphics[width = \linewidth,keepaspectratio,trim=0 0 0 48,clip]{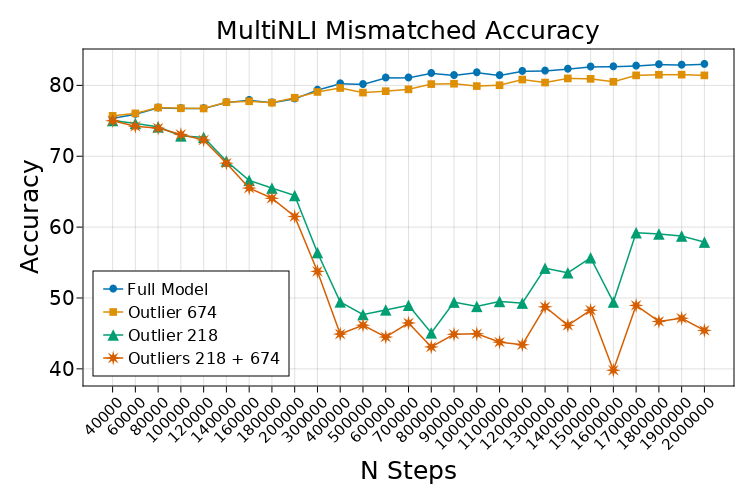}
    \caption{
        The accuracy on MNLI-mismatched of the checkpoints for BERT-base (seed 1), provided by \citet{sellam2022the-multiberts}. All classification heads are equally initialized.
    }
    \label{fig:fixed_seed_multiberts_seed_1_mnli_mm_acc_alone}
\end{figure}

%% file: figures/bert-base-uncased/bert-base-uncased_pos_shift.tex
\begin{figure*}[!ht]
    \centering
    \subcaptionbox{\label{fig:bert-base-uncased_outlier_pos_shift_full_model} Full model}[0.49\linewidth]{
        \adjustimage{max size={0.95\linewidth}{\textheight}}{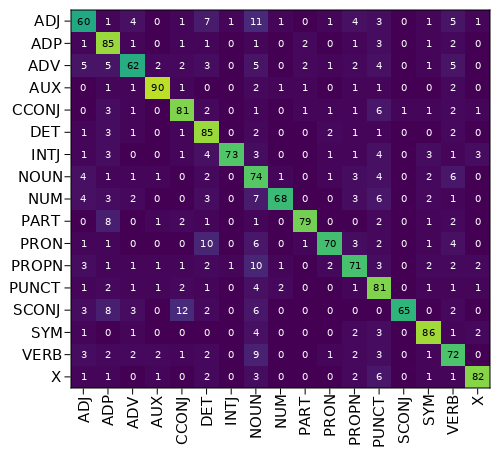}
    }
    \subcaptionbox{\label{fig:bert-base-uncased_outlier_pos_shift_308} Outlier 308 removed}[0.49\linewidth] {
        \adjustimage{max size={0.95\linewidth}{\textheight}}{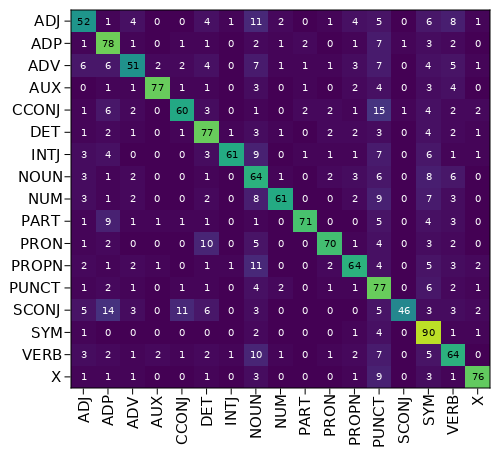}
    }
    \subcaptionbox{\label{fig:bert-base-uncased_outlier_pos_shift_381} Outlier 381 removed}[0.49\linewidth] {
        \adjustimage{max size={0.95\linewidth}{\textheight}}{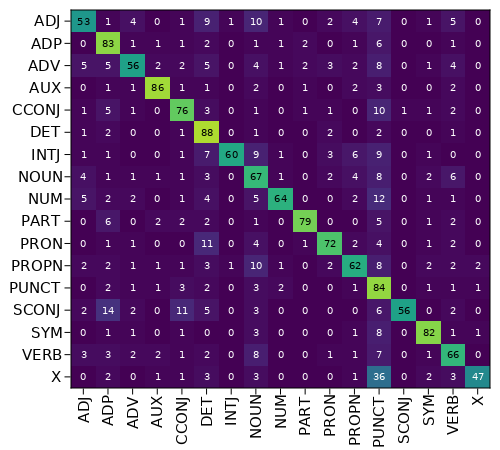}
    }
    \subcaptionbox{\label{fig:bert-base-uncased_outlier_pos_shift_308_381}Outliers 308 and 381 removed}[0.49\linewidth] {
        \adjustimage{max size={0.95\linewidth}{\textheight}}{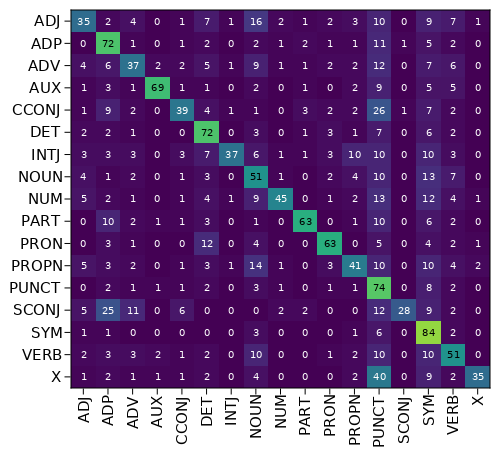}
    }
    \caption{
        The shift in percentage between the POS tags generated through MLM for full BERT-base model(\subref{fig:bert-base-uncased_outlier_pos_shift_full_model})
        and with different outliers removed, number 308 (\subref{fig:bert-base-uncased_outlier_pos_shift_308}), number 381 (\subref{fig:bert-base-uncased_outlier_pos_shift_381}) and together number 308 an number 381 (\subref{fig:bert-base-uncased_outlier_pos_shift_308_381}).
    }
    \label{fig:bert-base-uncased_outlier_pos_shift}
\end{figure*}

%% file: figures/bert-base-uncased/fine_tuned_bert-base-uncased_correlation_with_frequency.tex
\begin{figure}[!h]
    \centering
    \subcaptionbox{\label{fig:fine_tuned_outlier_corrtofreq_bert-base-uncased_w_special_tok} With special tokens}[0.49\linewidth]{
        \includegraphics[width = \linewidth,keepaspectratio]{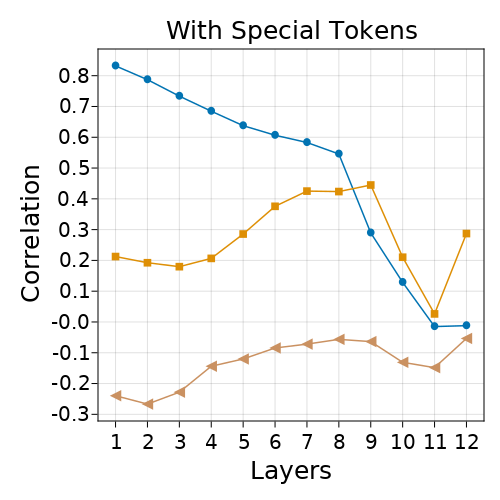}
    }
    \subcaptionbox{\label{fig:fine_tuned_outlier_corrtofreq_bert-base-uncased_wo_special_tok}Without special tokens}[0.49\linewidth]{
        \includegraphics[width = \linewidth,keepaspectratio]{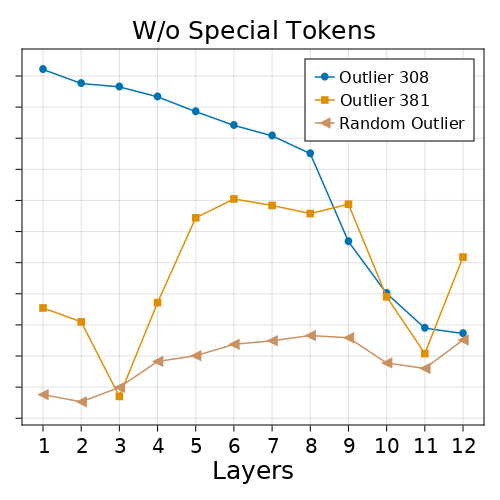}
    }
    \caption{
        BERT-base (fine tuned on MNLI) encoding Wikitext-v2 validation set data: the correlation between magnitude of hidden state parameters corresponding to outlier dimensions, and frequency of encoded tokens in pre-training data.
    }
    \label{fig:fine_tuned_outlier_corrtofreq_bert-base-uncased}
\end{figure}

%% file: figures/bert-base-uncased/fine_tuned_bert-base-uncased_correlation_with_attention.tex
\begin{figure*}[!ht]
    \centering
    \subcaptionbox{\label{fig:fine_tuned_bert_corr_with_att_308_with_spec_toks} \out{308}}[0.32\linewidth]{
        \includegraphics[width=\linewidth,keepaspectratio]{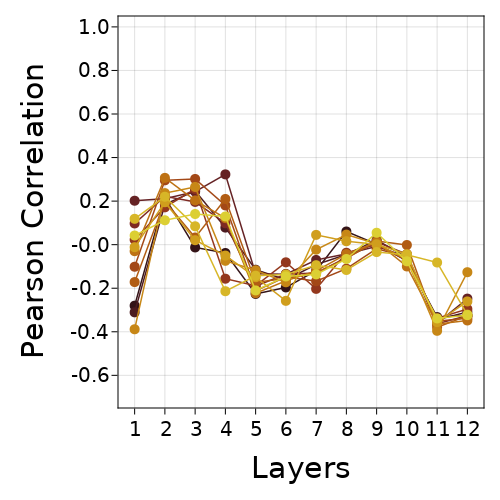}
    }
    \subcaptionbox{\label{fig:fine_tuned_bert_corr_with_att_381_with_spec_toks} \out{381}}[0.32\linewidth]{
        \figuretitle{Including special tokena}
        \includegraphics[width = \linewidth,keepaspectratio]{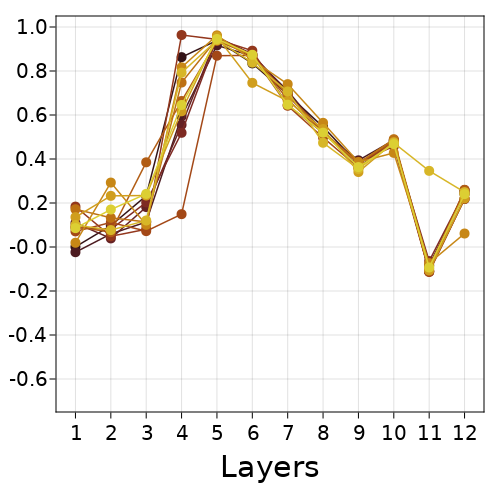}
    }
    \subcaptionbox{\label{fig:fine_tuned_bert_corr_with_att_random_with_spec_toks} Random coefficients}[0.32\linewidth]{
        \includegraphics[width = \linewidth,keepaspectratio]{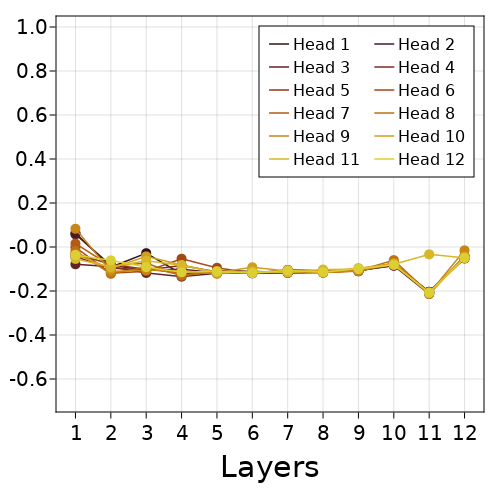}
    }
    \subcaptionbox{\label{fig:fine_tuned_bert_corr_with_att_308_without_spec_toks} \out{308}}[0.32\linewidth]{
        \includegraphics[width = \linewidth,keepaspectratio]{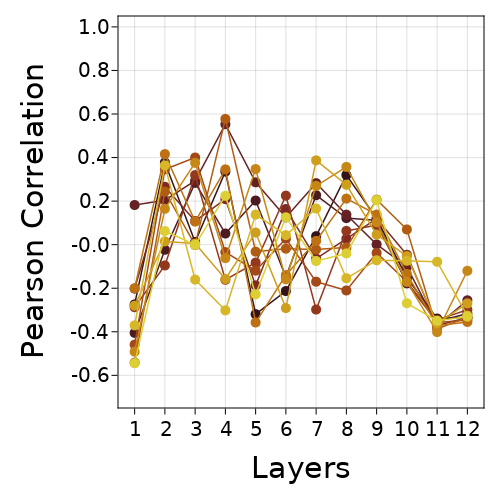}
    }
    \subcaptionbox{\label{fig:fine_tuned_bert_corr_with_att_381_without_spec_toks} \out{381}}[0.32\linewidth]{
        \vspace{7pt}
        \figuretitle{Excluding special tokena}
        \includegraphics[width = \linewidth,keepaspectratio]{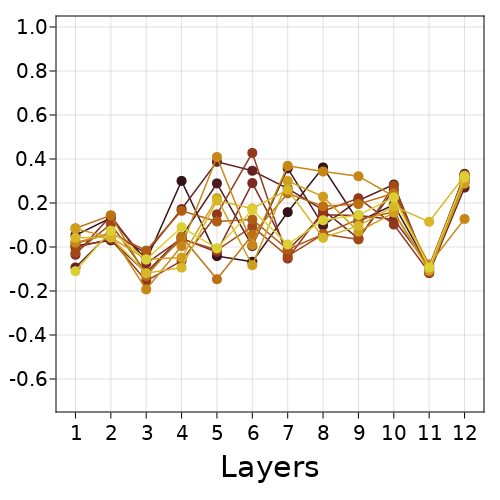}
    }
    \subcaptionbox{\label{fig:fine_tuned_bert_corr_with_att_random_without_spec_toks} Random coefficients}[0.32\linewidth]{
        \includegraphics[width = \linewidth,keepaspectratio]{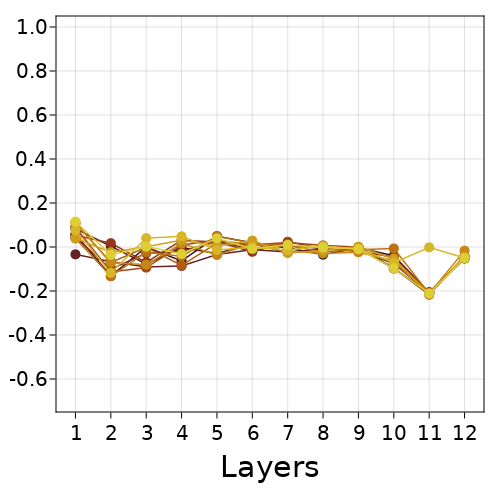}
    }
    
    \caption{Each figure shows the correlation between the \emph{average vertical attention values} in a BERT-base fine tuned on MNLI self-attention heads, and the magnitude of  hidden state parameters at the dimensions corresponding to outlier dimensions. The correlation is computed over examples from Wikitext-v2. Figures (\subref{fig:pre_trained_bert_corr_with_att_random_with_spec_toks}) and (\subref{fig:pre_trained_bert_corr_with_att_random_without_spec_toks}) show the average over 10 random dimensions.}
    \label{fig:fine_tuned_bert_corr_with_att}
\end{figure*}